\documentclass[sn-mathphys-num]{sn-jnl}
\usepackage{graphicx}%
\usepackage{multirow}%
\usepackage[font=small,labelfont=bf]{caption}
\usepackage{amsmath,amssymb,amsfonts}%
\usepackage{amsthm}%
\usepackage{mathrsfs}%
\usepackage{comment}
\usepackage{float}
\usepackage[T1]{fontenc}
\usepackage[title]{appendix}%
\usepackage{xcolor}%
\usepackage{amsmath}
\usepackage{amssymb}
\usepackage{mathptmx}  
\usepackage{caption}
\usepackage{float}
\restylefloat{figure}
\usepackage{textcomp}%
\usepackage{url}
\usepackage{caption}
\usepackage{subfigure}
\usepackage{manyfoot}%
\usepackage{booktabs}%
\usepackage{algorithm}%
\usepackage{algorithmicx}%
\usepackage{algpseudocode}%
\usepackage{listings}%
\usepackage{changepage}
\begin{document}

\title[Article Title]{A Comprehensive Methodological Survey of Human
Activity Recognition Across Divers Data Modalities}

\author*[1]{\fnm{Jungpil} \sur{Shin}}\email{jpshin@u-aizu.ac.jp}
\equalcont{These authors contributed equally to this work.}
\author[1]{\fnm{Najmul} \sur{Hassan}}
\equalcont{These authors contributed equally to this work.}
\author[1]{\fnm{Abu Saleh Musa} \sur{Miah}}
\equalcont{These authors contributed equally to this work.}
\author[1]{\fnm{Satoshi } \sur{Nishimura}}
\equalcont{These authors contributed equally to this work.}

\affil*[1]{\orgdiv{School of Computer Science and Engineering}, \orgname{The University of Aizu}, \orgaddress{\city{Aizuwakamatsu}, \state{Fukushima}, \country{Japan}}}

\abstract
{Human Activity Recognition (HAR) systems aim to understand human behaviour and assign a label to each action, attracting significant attention in computer vision due to their wide range of applications. HAR can leverage various data modalities, such as RGB images and video, skeleton, depth, infrared, point cloud, event stream, audio, acceleration, and radar signals. Each modality provides unique and complementary information suited to different application scenarios. Consequently, numerous studies have investigated diverse approaches for HAR using these modalities. This paper presents a comprehensive survey of the latest advancements in HAR from 2014 to 2024, focusing on machine learning (ML) and deep learning (DL) approaches categorized by input data modalities. We review both single-modality and multi-modality techniques, highlighting fusion-based and co-learning frameworks. Additionally, we cover advancements in hand-crafted action features, methods for recognizing human-object interactions, and activity detection. Our survey includes a detailed dataset description for each modality and a summary of the latest HAR systems, offering comparative results on benchmark datasets. Finally, we provide insightful observations and propose effective future research directions in HAR.}

\maketitle

\section{Introduction}
\label{sec1}
Human action recognition (HAR) has been a very active research topic for the past two decades in the field of computer vision and artificial intelligence (AI). That focuses on the automated analysis and understanding of human actions and recognition based on the movements and poses of the entire body. HAR plays an important role in various applications such as surveillance; healthcare \cite{papadopoulos2014real,presti20163d,islam2024multilingual,rahim2020hand}, remote monitoring, intelligent human-machine interfaces, entertainment, storage video, and retrieval \cite{van2015apt,zhu2017tornado} human-computer interaction\cite{lara2012survey,ziaeefard2015semantic,kibria2020creation}.\\
However, monitoring in 24 hours for security purposes makes it difficult to detect HAR. HAR is very important in computer vision and covers many research topics, including HAR in video, human tracking, and analysis and understanding in videos captured with a moving camera, where motion patterns exist due to video objects and moving camera as well \cite{wu2011action}. In such a scenario, it becomes ambiguous to recognize objects. The HAR methods were categorized into three distinct tiers: human action detection, human action tracking, and behaviour understanding methods.
 In recent years, the investigation of interaction \cite{herath2017going,chao2015hico,le2014tuhoi}  and human action detection\cite{peng2016multi,liu2018multi,patrona2018motion} has emerged as a prominent area of research.  
Many state-of-the-art techniques deal with action recognition using action frames as images and are only able to detect the presence of an object in it. They cannot properly recognize the object in an image or video. By properly recognizing an action in a video, it is possible to recognize the class of action more accurately. To perform action recognition, there has been an increased interest in this field in recent years due to the increased availability of computing resources as well as new advances in ML \cite{bengio2013representation} and DL.
Robust human action modelling and feature representation are essential components for achieving effective HAR. The main issue of representing and selecting features is a well-established problem within the fields of computer vision and ML \cite{bengio2013representation}.
Unlike the representation of features in an image domain, the representation of features of human actions in a video not only depicts the visual attributes of the human being(s) within the image domain but also must the extraction of alterations in visual attributes and pose. The problem representation of features has been expanded from a 2D space to a 3D spatio-temporal context. In the past few years, many types of action representation techniques have been proposed. These techniques include various approaches, such as local and global features that rely on temporal and spatial alterations \cite{das2016comprehensive,nguyen2014stap,shao2013spatio}, trajectory features that are based on key point tracking \cite{burghouts2014instantaneous}, motion changes that are derived from depth information\cite{yang2014super,ye2013survey} and action features that are derived from human pose changes\cite{li2016human,yang2014effective}.
With the performance and successful application of DL to activity recognition and classification, many researchers have used DL for HAR. This facilitates the automatically learned features from the video data set\cite{simonyan2014two,tran2015learning}.  However, the aforementioned review articles have only examined certain specific facets, such as the spatial, temporal interesting point (STIP) and HOF-found techniques for HAR, as well as the approaches for analyzing human walking and DL-based techniques. Numerous novel approaches have been recently developed, primarily about the utilization of depth learning techniques for feature learning. Hence, a comprehensive examination of these fresh approaches for recognizing human actions is of significant interest.
 
 \subsection{Article Search and Survey Methodology}\label{sec3.1}
The first step in conducting a comprehensive literature review involves gathering all relevant documents from 2014 to 2024 for Human Activity Recognition (HAR). This entails a meticulous screening process, including downloading and scrutinizing materials related to science, technology, or computer science. Publications are broadly categorized into journals, proceedings, book chapters, and lecture notes, focusing on articles presenting in-depth analysis and commentary.
Initially, articles were collected using relevant keywords such as:

\begin{itemize}
    \item Human Action Recognition, Human Activity Recognition (HAR)
    \item Action features including RGB, Skeleton, Sensor, Multimodality datasets
    \item ML and DL-based HAR
\end{itemize}
Subsequently, additional pertinent studies were incorporated after the initial selection of literature. Finally, supplementary investigations derived from the action recognition multimodal dataset were included to finalize this study.
In our investigation, most of the literature was collected from scholarly periodicals, journals, and conferences on computer vision. We prioritized articles published in prestigious journals and conferences such as:
\begin{itemize}
    \item IEEE Transactions on Pattern Analysis and Machine Intelligence (TPAMI)
    \item IEEE Transactions on Image Processing (TIP)
    \item International Conference on Computer Vision and Pattern Recognition (CVPR)
    \item IEEE International Conference on Computer Vision (ICCV)
    \item Springer, ELSEVIER, MDPI, Frontier, etc.
\end{itemize}
Simultaneously, to ensure our paper includes comprehensive methodologies, we selectively adopted a fundamental or exemplar approach when discussing similar methods in detail.
\subsubsection{Inclusion and Exclusion Criteria}
To refine and ensure relevance in our initial search results, we applied the following criteria:\\
\textbf{Inclusion Criteria:}
\begin{itemize}
    \item Publication date between 2014 and 2024;
    \item Inclusion of journals, proceedings, book chapters, and lecture notes;
    \item Focus on RGB-based, skeleton-based, sensor-based, and fusion HAR methods;
    \item Emphasis on the evolution of data acquisition, environments, and human activity portrayals.
\end{itemize}
\textbf{Exclusion Criteria:}
\begin{itemize}
    \item Exclusion of studies lacking in-depth information about their experimental procedures;
    \item Exclusion of research articles where the complete text isn't accessible, both in physical and digital formats;
    \item Exclusion of research articles that include opinions, keynote speeches, discussions, editorials, tutorials, remarks, introductions, viewpoints, and slide presentations.
\end{itemize}
\subsubsection{Article Selection}
\begin{figure}[ht]
    \centering
    \begin{adjustwidth}{-2cm}{1cm}
        \includegraphics[scale=0.50]{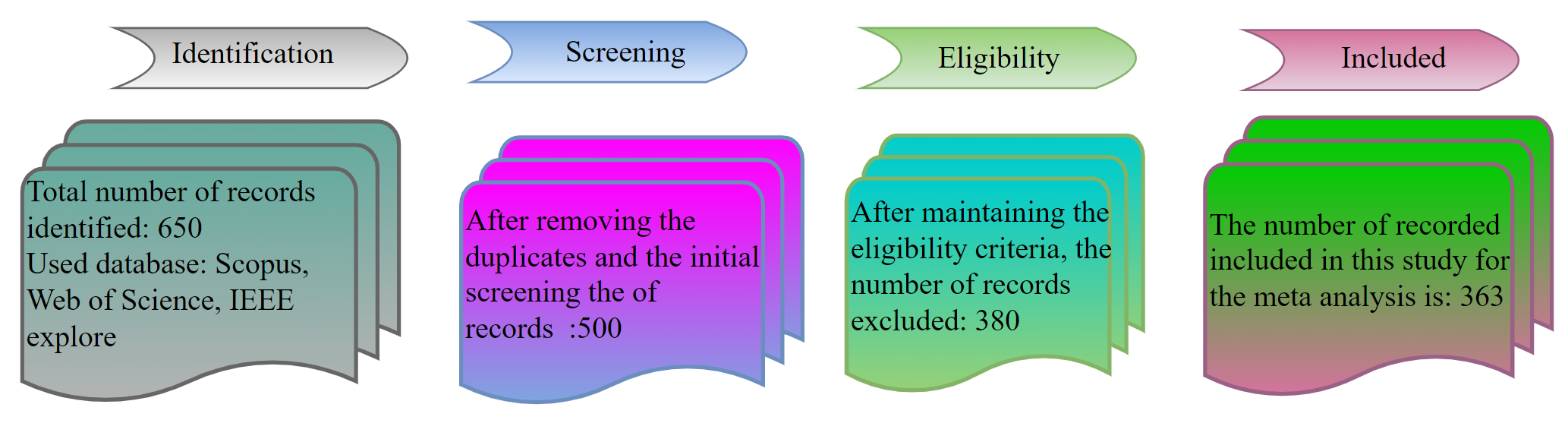}
    \caption{Article selection process block diagram.}
    \label{fig:article_selection}
      \end{adjustwidth}
\end{figure}

\begin{figure}[ht]
    \centering
        \includegraphics[scale=0.60]{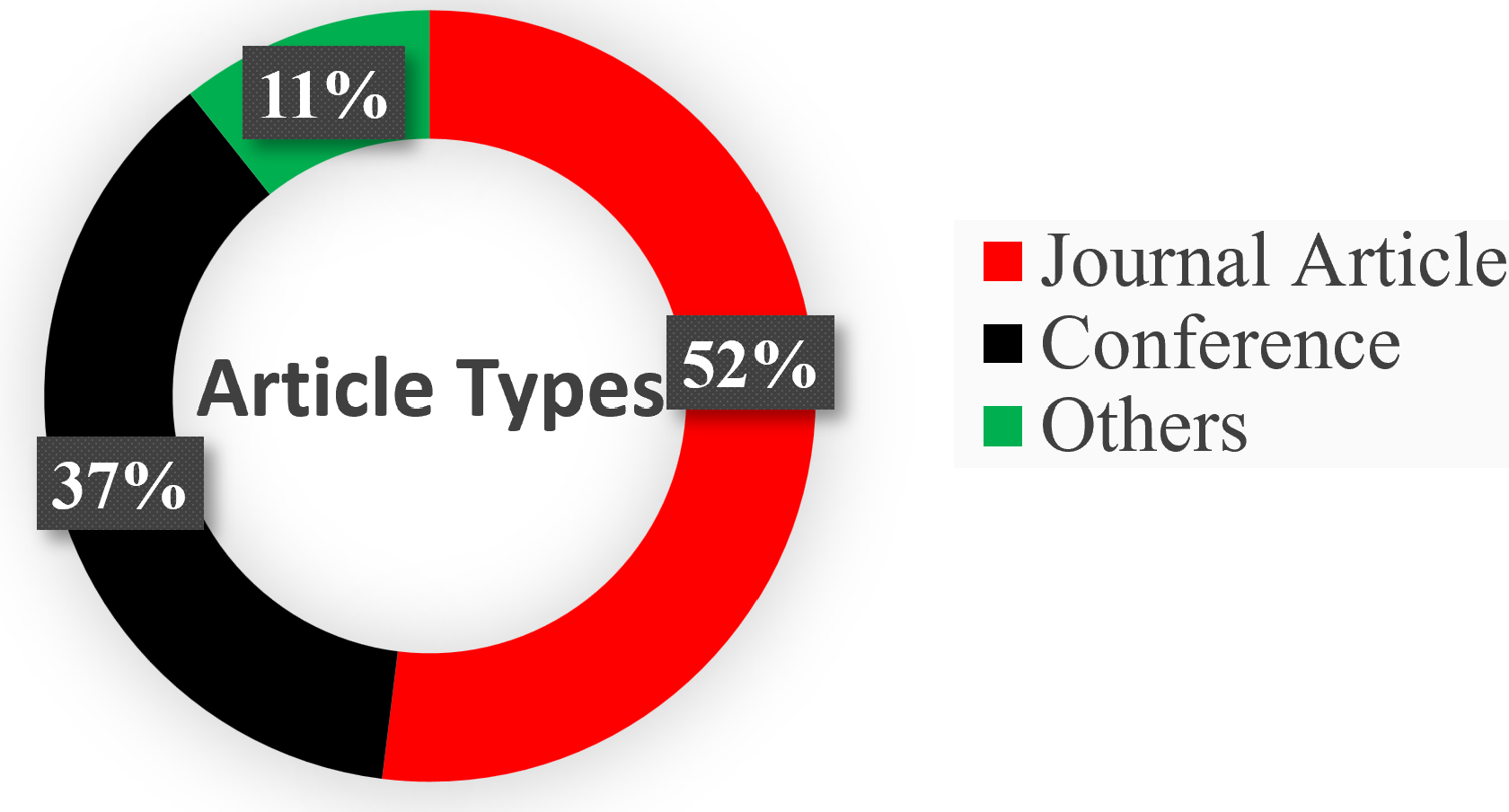}
    \caption{Article types journal conferences and others.}
    \label{fig:percentage_vise}
\end{figure}

\begin{figure}[ht]
    \centering
        \includegraphics[scale=0.60]{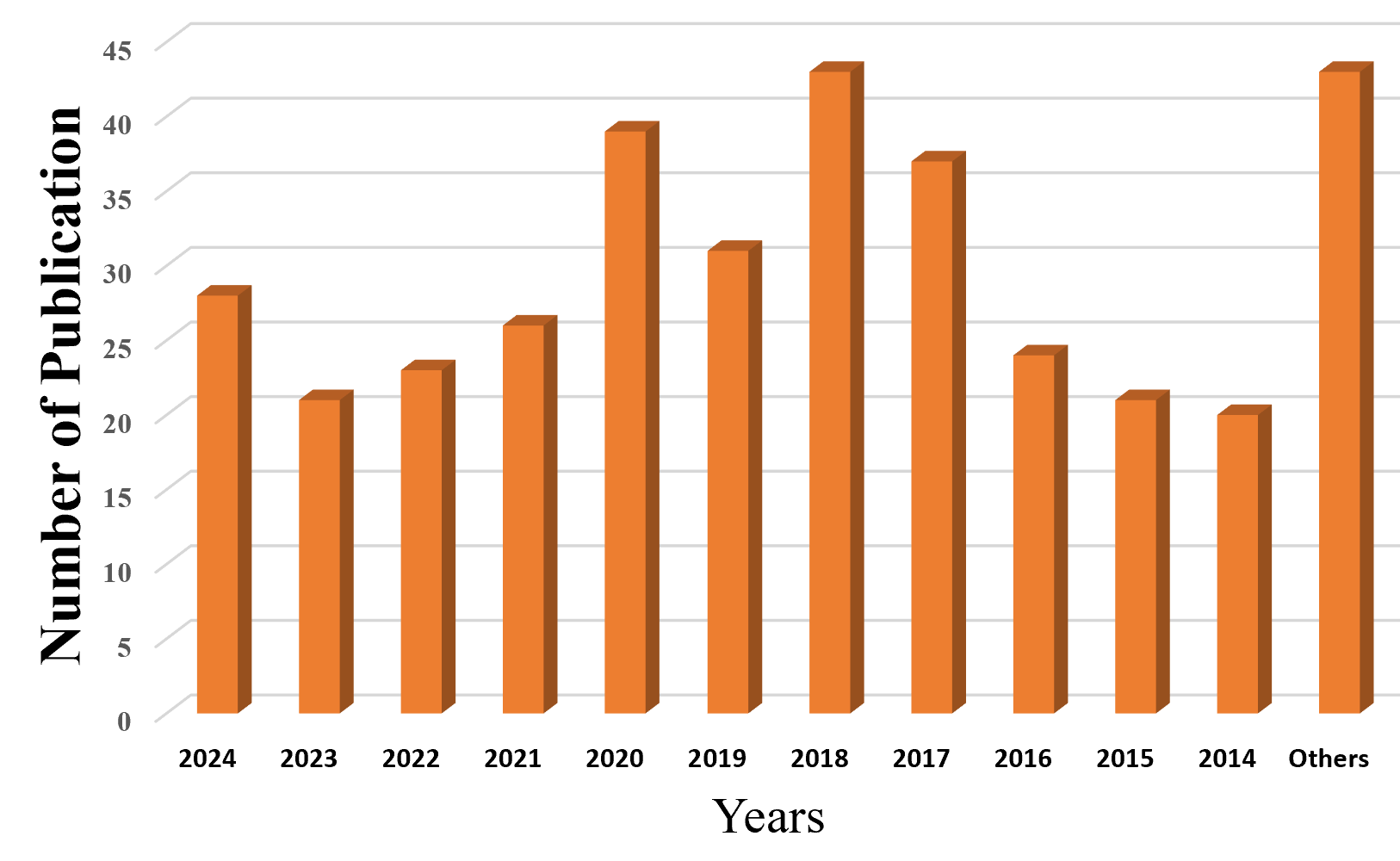}
    \caption{Yearwise papers gathering in this study.}
    \label{fig:yers_vise}
\end{figure}

We conducted a thorough survey of HAR methods, focusing on the evolution of data acquisition, environments, and human activity portrayals from 2014 to 2024. The preference is given to articles published in prestigious journals and conferences.
Figure \ref{fig:article_selection}  depicts the article selection process, illustrating the systematic approach adopted. Figure \ref{fig:percentage_vise} demonstrates the percentage of the journal, conference, and other ratios.
Figure \ref{fig:yers_vise} shows the year-wise number of references.

\subsubsection{Keywords and Search Strategy}
Two primary keywords, ``HAR'' and ``computer vision,'' determine the study's focal point. These keywords, supplemented by additional relevant terms, form the backbone of our search strategy across various databases and resources. Various materials, including original articles, review articles, book chapters, conference papers, and lecture notes, were gathered to review the subject matter comprehensively.
We reviewed each article through a structured process involving:

\begin{itemize}
    \item Abstract review
    \item Methodology analysis
    \item Discussion
    \item Result evaluations
\end{itemize}
Different modalities used in HAR have unique features, each with advantages and disadvantages in various tables. This approach ensures a thorough and systematic review of the HAR literature, providing a solid foundation for understanding the advancements and trends in this field.


\subsection{Motivation}
Many researchers have been working to develop a HAR system using various technologies, including ML and DL techniques with diverse feature extraction techniques. Herath et al.\cite{herath2017going} reported the classification techniques specific to HAR, disregarding an inquiry into the methods of interaction recognition and detecting actions. It is worth noting that in a recent study, Yu et al. \cite{yu1806human} performed a comprehensive analysis of the existing literature on the topic of action recognition and action prediction. In addition, the literature was summarized by \cite{wang2019deep}  within the framework of three key areas: sensor modality, deep models, and application.
Guo et al. \cite{presti20163d} analyzed methods employed in human HAR with still images, exploring various ML and  DL-based approaches for extracting low-level features and representing actions at higher levels. Vrigkas et al. \cite{vrigkas2015review} also reviewed HAR using RGB static images, covering both single-mode and multi-mode approaches. Vishwakarma et al. \cite{vishwakarma2013survey} summarized classical HAR methods, categorizing them into hierarchical and non-hierarchical methods based on feature representation. The survey by Ke et al. \cite{ke2013review} provided a comprehensive overview of handcrafted methods in HAR. Additionally, surveys \cite{zhu2020comprehensive}, \cite{zhang2019comprehensive}, \cite{kong2022human}, \cite{ma2022survey} extensively discuss the strengths and weaknesses of handcrafted versus DL methods, emphasizing the advantages of DL-based approaches. Xing et al. \cite{xing2021deep} focused on HAR development using 3D skeleton data, reviewing various DL-based techniques and comparing their performance across different dimensions.
Presti et al. \cite{presti20163d} presented HAR techniques based on 3D skeleton data. Methods for HAR using depth and skeleton data have been thoroughly reviewed by Ye et al. \cite{ye2013survey}; they also present HAR techniques using depth data. \\
 Although certain review articles discuss data fusion methods, they offer a limited overview of HAR approaches to particular data types. Similarly, Subetha et al.\cite{subetha2016survey}  presented the same strategy to review action recognition methods.  However, in distinction to those studies, we categorize HAR into four distinct categories: action recognition RGB and handcrafted features, action recognition RGB and DL, action recognition skeleton and handcrafted features, action recognition skeleton-based and DL, and action recognition using multimodal dataset. The crucial element of the analysis regarding the literature on HAR is that most surveys have focused on the representations of human action features. The data of the image sequences that have been processed are typically well-segmented and consist solely of a single action event. 
More recently, many researchers have been working to make an HAR survey study with some specific point of view. Such as some researchers surveyed graph convolutional network (GCN) structures
and data modalities for HAR and the application of GCNs in HAR \cite{feng2022skeleton},
\cite{feng2022comparative}.
Gupta et al. \cite{gupta2021quo} explored current and future directions in skeleton-based HAR and introduced the skeleton-152 dataset, marking a significant advancement in the field. Meanwhile, Song et al. \cite{song2021human} reviewed advancements in human pose estimation and its applications in HAR, emphasizing its importance. Additionally, Shaikh et al. \cite{shaikh2021rgb} focused on data integration and recognition approaches within a visual framework, specifically from an RGB-D perspective. Majumder et al. \cite{majumder2020vision} and Wang et al. \cite{wang2019comparative} provided reviews of popular methods using vision and inertial sensors for HAR. More recently, want et al. \cite{wang2023comprehensive} survey HAR by performing two modalities of RGB-based and skeleton-based HAR techniques. Similarly, Sun et al. \cite{sun2022human} survey  HAR with various multi-modality methods.

\subsection{Research Gaps and New Research Challenges}
Also, each survey paper can give us an overall summary of the existing work in this domain. Still, it lacks comparative studies of the RGB, Skeleton, sensor, and fusion-based diverse modality-based HAR system of the recent technologies. 
From a data perspective, most reviews on HAR are limited to methodologies based on specific data, such as RGB, depth, and fusion data modalities. Moreover, we did not find a HAR survey paper that included diverse modality-based HAR, including their benchmark dataset and latest performance accuracy for 2014-2024. The study inspires us \cite{wang2023comprehensive,herath2017going} to complete a survey study with current research trends for HAR. 

\subsection{Our Contribution}
\begin{figure*}[ht]
    \centering
    \includegraphics[scale=0.30]{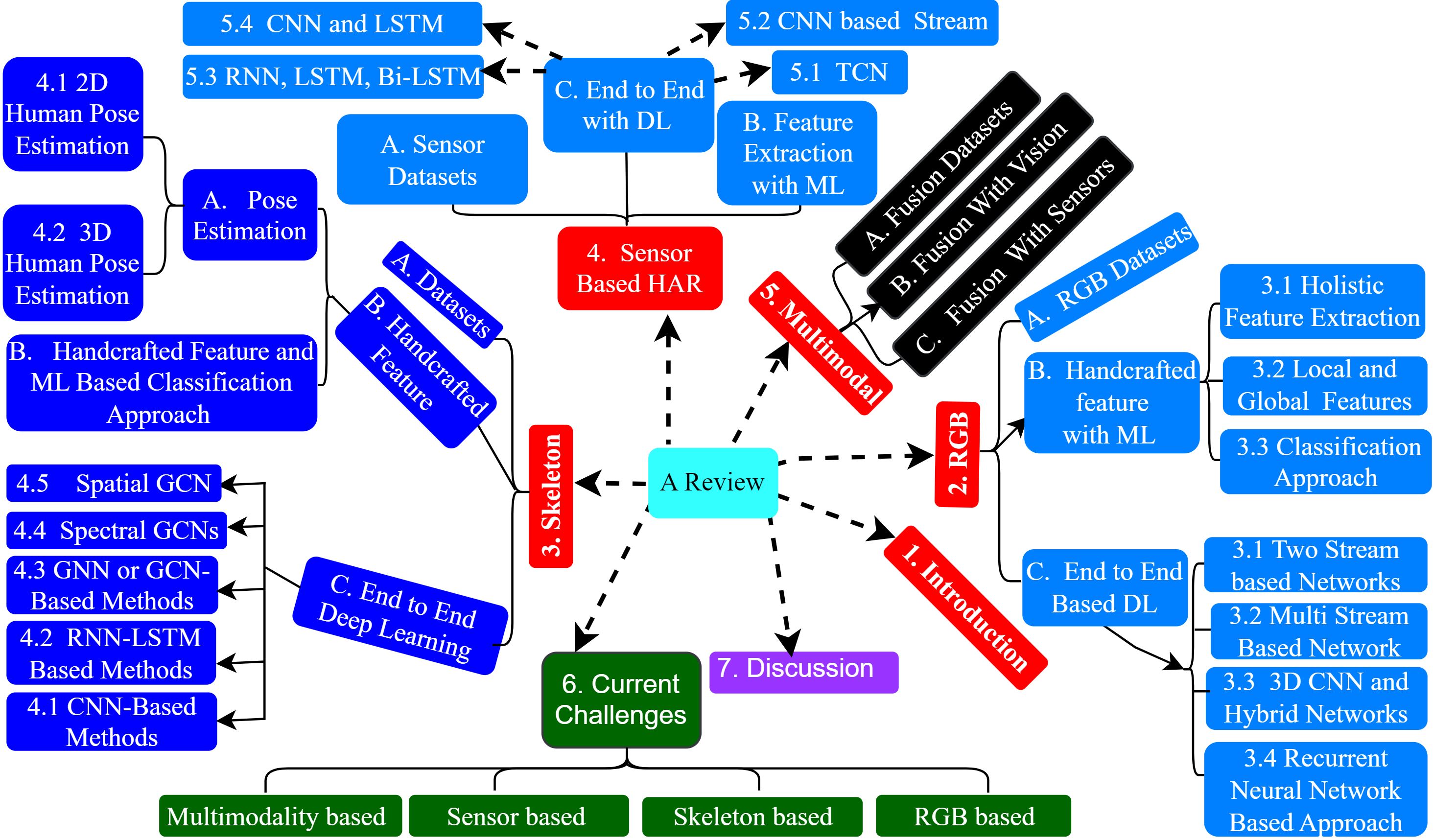}
    \caption{The structure of this paper.}
    \label{fig1_structure}
\end{figure*}

Figure \ref{fig1_structure} demonstrates the proposed methodology flowchart. In this study, we survey state-of-the-art methods for HAR, addressing their challenges and future directions across vision, sensor, and fusion-based data modalities. We also summarize the current 2 dimensions and 3 dimensions pose estimation algorithms before discussing skeleton-based feature representation methods. Additionally, we categorize action recognition techniques into handcrafted feature-based ML and end-to-end DL-based methods. Our main contributions are as follows:

\begin{itemize}
    \item 
    \textbf{Comprehensive Review with Diverse Modality}: We conduct a thorough survey of RGB-based, skeleton-based, sensor-based, and fusion HAR-based methods, focusing on the evolution of data acquisition, environments, and human activity portrayals from 2014 to 2024.
    \item 
    \textbf{Dataset Description}: We provide a detailed overview of benchmark public datasets for RGB, skeleton, sensor, and fusion data, highlighting their latest performance accuracy with reference. 
    \item 
    \textbf{Unique Process}: Our study covers feature representation methods, common datasets, challenges, and future directions, emphasizing the extraction of distinguishable action features from video data despite environmental and hardware limitations.
    \item 
    \textbf{Identification of Gaps and Future Directions}: We identify significant gaps in current research and propose future research directions supported by the latest performance data for each modality.
    \item 
    \textbf{Evaluation of System Efficacy}: We assess existing HAR systems by analyzing their recognition accuracy and providing benchmark datasets for future development.
    \item 
    \textbf{Guidance for Practitioners}: Our review offers practical guidance for developing robust and accurate HAR systems, providing insights into current techniques, highlighting challenges, and suggesting future research directions to advance HAR system development.
\end{itemize}

\subsection{Research Questions}
This research addresses the following major questions:
1. What are the main difficulties faced in Human activity recognition? \\
2. What are some challenges faced with Human activity recognition? \\
3. What are the major algorithms involved in Human activity recognition? \\


\subsection{Organization of the Work} 
The paper is categorized as follows. The benchmark datasets are provided in section \ref{section_1}. The action recognition RGB-data modality methods and skeleton data modality-based are discussed in sections \ref{section_2} and \ref{section_3}, respectively. In sections \ref{section_4}, \ref{section_5}, and \ref{section_6}, we introduce sensor modality-based human activity recognition, multimodal fusion modality-based, and current challenges, including four data modalities, respectively. We discuss future research trends and direction in sections \ref{section_7}. Finally, in the last section \ref{section_8}, we summarized the conclusions. The detailed structure of this paper is shown in Figure \ref{fig1_structure}.

\section{RGB-Data Modality Based Action Recognition Methods} \label{section_2}
\begin{figure}[ht]
    \centering
    \includegraphics[scale=0.25]{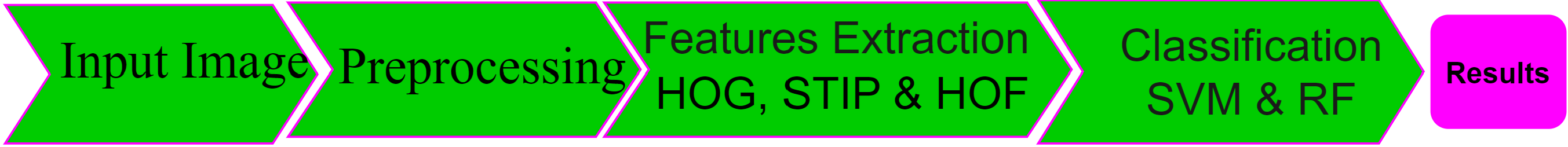}
    \caption{Action recognition RGB data and handcrafted features.}
    \label{fig1_frame}
\end{figure}

Figure \ref{fig1_frame} demonstrated a common workflow diagram of the RGB-based action recognition methods. The early stages of research about the HAR were conducted based on the RGB data, and initially, feature extraction mostly depended on manual annotation \cite{ullah2019action,lan2015beyond}. These annotations often relied on existing knowledge and prior assumptions. After this, DL-based architectures were developed to extract the most effective features and the best performances. The following sections describe the dataset, the methodological review of RGB-based handcrafted features with ML, and various ideas for DL-based approaches. Moreover, Table \ref{Tab:RGB_based} lists detailed information about the RGB data modality, including the datasets, features extraction methods, classifier, years, and performance accuracy. 

\subsection{RGB-Based Datasets of HAR} \label{section_1}
We provided the most popular benchmark HAR datasets, which come from the RGB skeleton, which is demonstrated in Table \ref{tab1:rgb_skeleton_dataset}. The dataset table demonstrated the details of the datasets, including modalities, creation year, number of classes, number of subjects who participated in recording the dataset, number of samples, and latest performance accuracy of the dataset with citation. 

The RGB dataset encompasses several prominent benchmarks for Human Activity Recognition (HAR). Notably, the Activity Net dataset, introduced in 2015, comprises 203 activity classes and an extensive 27,208 samples, achieving an impressive accuracy of 94.7\% in recent evaluations \cite{caba2015activitynet} \cite{li2022uniformerv2}. The Kinetics-400 and Kinetics-700 datasets, from 2017 and 2019 respectively, include 400 and 700 classes with approximately 306,245 and 650,317 samples. These datasets are notable for their high accuracy rates of 92.1\% and 85.9\% \cite{kay2017kinetics} \cite{carreira2019short} \cite{wang2024internvideo2}. The AVA dataset, also from 2017, contains 80 classes and 437 samples, with a recorded accuracy of 83.0\% \cite{gu2018ava} \cite{sheng2018attention}. The EPIC Kitchen 55 dataset from 2018 offers a comprehensive view with 149 classes and 39,596 samples. The Moments in Time dataset, released in 2019, is one of the largest with 339 classes and around 1,000,000 samples, although it has a relatively lower accuracy of 51.2\% \cite{monfort2019moments} \cite{wang2024internvideo2}. Each dataset is instrumental for training and evaluating HAR models, providing diverse scenarios and activities.

\begin{table}[]
\centering
\caption{Benchmark datasets for HAR  RGB and Skeleton.}
\label{tab1:rgb_skeleton_dataset}
\begin{tabular}{|c|c|c|c|c|c|c|c|}
\hline
\textbf{Dataset} & \textbf{Data set modalities} & \textbf{Year} & \textbf{Class} & \textbf{Subject} & \textbf{Sample} & \textbf{Latest Accuracy} \\ \hline
\begin{tabular}[c]{@{}c@{}}UPCV \cite{theodorakopoulos2014pose}\end{tabular} & Skeleton & 2014 & 10 & 20 & 400 & 99.20\% \cite{zhou2022high} \\ \hline
\begin{tabular}[c]{@{}c@{}}Activity Net \cite{caba2015activitynet}\end{tabular} & RGB & 2015 & 203 & - & 27208 & 94.7\% \cite{li2022uniformerv2} \\ \hline
\begin{tabular}[c]{@{}c@{}}Kinetics-400 \cite{kay2017kinetics}\end{tabular} & RGB & 2017 & 400 & - & \~306245 & 92.1\% \cite{wang2024internvideo2} \\ \hline
\begin{tabular}[c]{@{}c@{}}AVA \cite{gu2018ava}\end{tabular} & RGB & 2017 & 80 & - & 437 & 83.0\% \cite{sheng2018attention} \\ \hline
\begin{tabular}[c]{@{}c@{}}EPIC Kitchen 55 \cite{damen2018scaling}\end{tabular} & RGB & 2018 & 149 & 32 & 39596 & - \\ \hline
\begin{tabular}[c]{@{}c@{}}AVE \cite{tian2018audio}\end{tabular} & RGB & 2018 & 28 & - & 4143 & - \\ \hline
\begin{tabular}[c]{@{}c@{}}Moments in Times \cite{monfort2019moments}\end{tabular} & RGB & 2019 & 339 & - & \~1000000 & 51.2\% \cite{wang2024internvideo2} \\ \hline
\begin{tabular}[c]{@{}c@{}}Kinetics-700 \cite{carreira2019short}\end{tabular} & RGB & 2019 & 700 & - & \~650317 & 85.9\% \cite{wang2024internvideo2} \\ \hline
\begin{tabular}[c]{@{}c@{}}RareAct \cite{miech2020rareact}\end{tabular} & RGB & 2020 & 122 & 905 & 2024 & 49.80\% \\ \hline
\begin{tabular}[c]{@{}c@{}}HiEve \cite{lin2020human}\end{tabular} & RGB, Skeleton & 2020 & - & - & - & 95.5\% \cite{duan2024abnormal} \\ \hline
\begin{tabular}[c]{@{}c@{}}UPCV \cite{theodorakopoulos2014pose}\end{tabular} & Skeleton & 2014 & 10 & 20 & 400 & 99.20\% \cite{zhou2022high} \\ \hline
 \begin{tabular}[c]{@{}c@{}}MSRDaily\\Activity3D \cite{wang2012mining}\end{tabular} & RGB, Skeleton & 2012 & 16 & 10 & 320 & 97.50\% \cite{shahroudy2017deep} \\ \hline
\begin{tabular}[c]{@{}c@{}}N-UCLA \cite{wang2014cross}\end{tabular} & RGB, Skeleton& 2014 & 10 & 10 & 1475 & 99.10\% \cite{cheng2024dense} \\ \hline
\begin{tabular}[c]{@{}c@{}}Multi-View TJU \cite{liu2014multiple}\end{tabular} & RGB, Skeleton & 2014 & 20 & 22 & 7040 & - \\ \hline
\begin{tabular}[c]{@{}c@{}}UTD-MHAD \cite{chen2015utd}\end{tabular} & RGB, Skeleton & 2015 & 27 & 8 & 861 & 95.0\% \cite{liu2018recognizing} \\ \hline
\begin{tabular}[c]{@{}c@{}}UWA3D\\Multiview II  \cite{rahmani2016histogram}\end{tabular} & RGB, Skeleton &2015& 30 & 10 & 1075 & - \\ \hline
\begin{tabular}[c]{@{}c@{}}NTU RGB+D 60 \cite{shahroudy2016ntu}\end{tabular} & RGB, Skeleton & 2016 & 60 & 40 & 56880 & 97.40\% \cite{cheng2024dense} \\ \hline
\begin{tabular}[c]{@{}c@{}}PKU-MMD \cite{liu2017pku}\end{tabular} & RGB, Skeleton & 2017 & 51 & 66 & 10076 & 94.40\% \cite{li2019making} \\ \hline
\begin{tabular}[c]{@{}c@{}}NEU-UB \cite{kong2017max}\end{tabular} & RGB & 2017 & 6 & 20 & 600 & - \\ \hline
\begin{tabular}[c]{@{}c@{}}Kinetics-600 \cite{carreira2018short}\end{tabular} & RGB, Skeleton& 2018 & 600 & - & 595445 & 91.90\% \cite{wang2024internvideo2} \\ \hline
\begin{tabular}[c]{@{}c@{}}RGB-D \\Varing-View \cite{ji2018large}\end{tabular} & RGB, Skeleton & 2018 & 40 & 118 & 25600 & - \\ \hline
\begin{tabular}[c]{@{}c@{}}NTU RGB+D 120 \cite{liu2019ntu}\end{tabular} & RGB, Skeleton & 2019 & 120 & 106 & 114480 & 95.60\% \cite{cheng2024dense} \\ \hline
\begin{tabular}[c]{@{}c@{}}Drive\&Act \cite{martin2019drive}\end{tabular} & RGB, Skeleton & 2019 & 83 & 15 & - & 77.61\% \cite{lin2024multi} \\ \hline
\begin{tabular}[c]{@{}c@{}}MMAct \cite{kong2019mmact}\end{tabular} & RGB, Skeleton & 2019 & 37 & 20 & 36764 & 98.60\% \cite{liu2021semantics} \\ \hline
\begin{tabular}[c]{@{}c@{}}Toyota-SH \cite{das2019toyota}\end{tabular} & RGB, Skeleton & 2019 & 31 & 18 & 16115 & - \\ \hline
\begin{tabular}[c]{@{}c@{}}IKEA ASM \cite{ben2021ikea}\end{tabular} & RGB, Skeleton & 2020 & 33 & 48 & 16764 & - \\ \hline
\begin{tabular}[c]{@{}c@{}}ETRI-Activity3D \cite{jang2020etri}\end{tabular} & RGB, Skeleton & 2020 & 55 & 100 & 112620 & 95.09\% \cite{dokkar2023convivit} \\ \hline
\begin{tabular}[c]{@{}c@{}}UAV-Human \cite{li2021uav}\end{tabular} & RGB, Skeleton & 2021 & 155 & 119 & 27428 & 55.00\% \cite{xian2024pmi} \\ \hline
\end{tabular}
\end{table}

\subsection{Handcrafted Features with ML-Based Approach}
Researchers employed handcrafted feature extraction with ML-based systems at early ages to develop HAR systems \cite{patel2018human}. In the action representation step, the RGB data is utilized to transform into the feature vector, and these feature vectors are fed into the classifier \cite{liu2011recognizing,shi2011human} to get the desired results of the action classification step. Table \ref{Tab:handcrafed_based} shows the analysis of the handcrafted-based approach, including the datasets, methods of feature extraction, classifier, years, and performance accuracy.
Handcrafted features are designed to capture the physical motions performed by humans and the spatial and temporal variations depicted in videos that portray actions. These variations include methods that utilize the spatiotemporal volume-based representation of actions, methods based on the Space-Time Interest Point (STIP), methods that rely on the trajectory of skeleton joints for action representation, and methods that utilize human image sequences for action representation. 
 Chen et al. \cite{chen2015action} demonstrate this by employing DMM-based gestures for motion information extraction, while Local Binary Pattern (LBP) feature encoding enhances discriminative power for action recognition. Meanwhile, Patel et al. \cite{patel2018human} fuse various features, including HOG and  LBP, to improve network performance in recognizing human activities. The handcrafted feature can be categorized as below: 

\subsubsection{Holistic Feature Extraction}
Many researchers have been working to develop Human Activity Recognition (HAR) systems based on holistic features and machine learning algorithms. Holistic representation aims to capture motion information of the entire human subject. Spatiotemporal action recognition often uses template-matching techniques, with key methods focusing on creating effective action templates.
Bobick et al. introduced two approaches, Motion Energy Image (MEI) and Motion History Image (MHI), to perform action representation \cite{bobick2001recognition}. Meanwhile, Zhang et al. utilized polar coordinates in MHI and developed a Motion Context Descriptor (MCD) based on the Scale-Invariant Feature Transform (SIFT) \cite{zhang2008motion}. Somasundaram et al. applied sparse representation and dictionary learning to calculate video self-similarity in both time and space \cite{somasundaram2014action}. In scenarios with a stationary camera, these approaches effectively capture shape-related information like human silhouettes and contours through background subtraction.
However, accurately capturing silhouettes and contours in complex scenes or with camera movements remains challenging, especially when the human body is partially obscured. Many methods employ a sliding window approach to detect multiple actions within the same scene, which can be computationally expensive. These approaches transform dynamic human motion into a holistic representation in a single image. While they capture relevant foreground information, they are sensitive to background noise, including irrelevant information.

\subsubsection{Local and Global Representation}
Holistic feature extraction techniques for HAR face several limitations, including sensitivity to background noise, reliance on stationary cameras, difficulty in complex scenes, occlusion issues, high computational cost, limited robustness to variations, and neglect of contextual information, making them less effective in dynamic, real-world scenarios.
\begin{figure*}[ht]
\begin{adjustwidth}{-2cm}{0cm}
  \centering
    \includegraphics[scale=0.55]{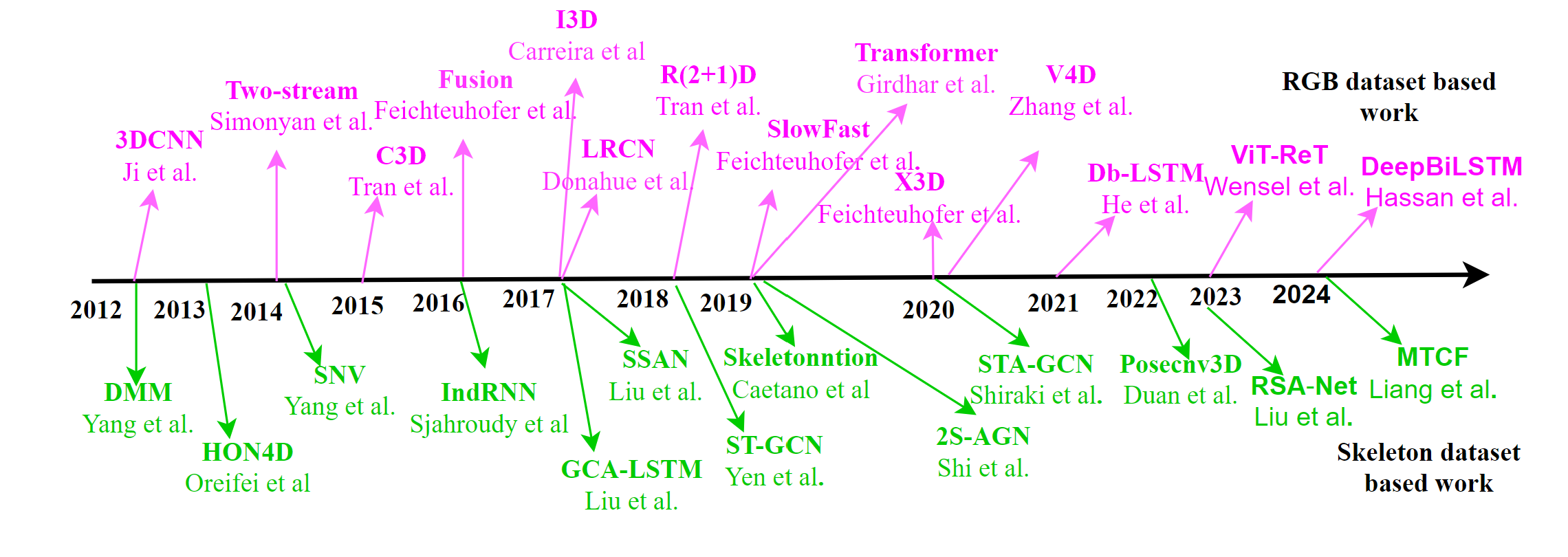}
    \caption{Milestone approaches for HAR: RGB-based milestone methods are in pink font, while skeleton-based milestone methods are in green font}
    \label{fig_milestone_RGB_skeleton}
   \end{adjustwidth}
\end{figure*}

Combining local and global representations can effectively address HAR's holistic feature extraction limitations. Local features reduce background noise sensitivity and handle occlusions, while global features ensure comprehensive activity recognition. This combination enhances robustness to variations, manages complex scenes, and optimizes computational efficiency, improving HAR accuracy and reliability.
 The local presentation means identifying a specific region, while the global representation means identifying the whole region with significant motion information.  These methods \cite{shao2013spatio,das2016comprehensive,nguyen2014stap} contain local and global features based on spatial-temporal changes trajectory attributes that are founded on key point tracking \cite{wang2013action,burghouts2014instantaneous}, motion changes that are derived from depth information\cite{yang2014super,ye2013survey,oreifej2013hon4d} and action-based features that are predicated on human pose changes \cite{li2016human,yang2014effective}.
The HoG is one of the feature-based techniques that calculate features on the base orientation of gradients in an image or video sequence. The HoG features are then used to encode local and global texture information, aiming to recognize different actions. Some of the presented approaches exploit the HoG in action recognition, including \cite{patel2020histogram,tan2020human,wattanapanich2021analysis,zuo2019histogram,sahoo2019fusion,ma2013action} in various ways. Histogram of optical flow (HOF) is a used feature extraction method in action recognition \cite{xia2021human,nebisoy2021video,wang2020enhanced,hassan2018temporal}. It involves building histograms to present different actions over the spatio-temporal domain in a video. However, in this method, the number of bins needs to be set in advance. The challenge addresses cluttered backgrounds and camera movement by performing a physical feature-driven approach HOF.
\subsubsection{Classification Approach}
Once we have the feature representation, we feed it into classifiers such as support vector machine (SVM) \cite{marszalek2009actions,laptev2008learning,schuldt2004recognizing}, random forest, and KNN \cite{blank2005actions,laptev2007retrieving,tran2008human} to predict the activity label. While some classification methods based on sequential such as Hidden Markov Models (HMM), Condition Random Fields (CRF) \cite{morency2007latent,wang2006hidden,wang2007recognizing}, Structured Support Vector Machine (SSVM) \cite{shi2011human,tang2012learning,wang2012substructure}, and Global Gaussian Mixture Models (GGMM) \cite{wu2011action}  these approaches perform sequential based for classification tasks. Additionally, luo et al. utilized features fusion-based methods, Maximum Margin Distance Learning (MMDL) \cite{luo2014learning} and Multi-task Spare Learning Model (MTSLM) \cite{yuan2013multi}. These methods perform the classification task based on combining various characteristics to enhance the classification task.

\begin{table}[]
\centering
\setlength{\tabcolsep}{2pt}
   \caption{Handcrafted features based on existing techniques for action recognition.}
\label{Tab:handcrafed_based}
\begin{tabular}{|c|c|c|c|c|c|c|}
\hline
\textbf{Author} & \textbf{Year} & \textbf{Dataset Name} & \textbf{Modality} & \textbf{Method} & \textbf{Classifier} & \textbf{Accuracy [\%]} \\ \hline
Chakraborty et al. \cite{chakraborty2011selective} & 2011 & \begin{tabular}{@{}c@{}}Weizmann\\KTH\\You Tube\end{tabular} & RGB & STIP & SVM & \begin{tabular}{@{}c@{}}100.00\\96.35\\86.98\end{tabular} \\ \hline
Gan et al. \cite{gan2013human} & 2013 & UTKinect-Action & RGB & RF & APJ3D & 92.00 \\ \hline
Everts et al. \cite{everts2014evaluation} & 2014 & \begin{tabular}{@{}c@{}}UCF11\\UCF50\end{tabular} & RGB & multi-channel STIP & SVM & \begin{tabular}{@{}c@{}}78.6\\72.9\end{tabular} \\ \hline
Zhu et al. \cite{zhu2014evaluating} & 2014 & \begin{tabular}{@{}c@{}}MSRAction3D\\UTKinectAction\\CAD-60\\MSRDailyActivity3D\\HMDB51\end{tabular} & RGB & STIP (HOG/HOF) & SVM & \begin{tabular}{@{}c@{}}94.3\\91.9\\87.5\\80.0\end{tabular} \\ \hline
Yang et al. \cite{yang2014effective} & 2014 & MSR Action3D & RGB & EigenJoints-based & NBNN & 97.8 \\ \hline
Liu et al. \cite{liu2015learning} & 2015 & \begin{tabular}{@{}c@{}}KTH\\HMDB51\\UCF YouTube\\Hollywood2\end{tabular} & RGB & GP-learned descriptors & SVM & \begin{tabular}{@{}c@{}}95.0\\48.4\\82.3\\46.8\end{tabular} \\ \hline
Xu et al. \cite{xu2016human} & 2016 & \begin{tabular}{@{}c@{}}MSRAction3D\\UTKinectAction\\Florence 3D-Action\end{tabular} & RGB & PSO-SVM & - & \begin{tabular}{@{}c@{}}93.75\\97.45\\91.20\end{tabular} \\ \hline
Vishwakarma et al. \cite{vishwakarma2016proposed} & 2016 & \begin{tabular}{@{}c@{}}KTH\\Weizmann\\i3Dpost\\Ballet\\IXMAS\end{tabular} & RGB & SDEG & SVM & \begin{tabular}{@{}c@{}}95.5\\100\\92.92\\93.25\\85.8\end{tabular} \\ \hline
Singh et al. \cite{singh2017graph} & 2017 & \begin{tabular}{@{}c@{}}UCSDped-1\\UCSDped-2\\UMN\end{tabular} & RGB & Graph formulation & SVM & \begin{tabular}{@{}c@{}}97.14\\90.13\\95.24\end{tabular} \\ \hline
Jalal et al. \cite{jalal2017robust} & 2017 & \begin{tabular}{@{}c@{}}IM-DailyDepthActivity\\MSRAction3D\\MSRDailyActivity3D\end{tabular} & RGB & HOG-DDS & HMM & \begin{tabular}{@{}c@{}}72.86\\93.3\\97.9\end{tabular} \\ \hline
Nazir et al. \cite{nazir2018evaluating} & 2018 & \begin{tabular}{@{}c@{}}KTH\\UCF Sports\\UCF11\\Hollywood\end{tabular} & RGB & D-STBoE & SVM & \begin{tabular}{@{}c@{}}91.82\\94.00\\94.00\\68.10\end{tabular} \\ \hline
Ullah et al. \cite{ullah2021weakly} & 2021 & \begin{tabular}{@{}c@{}}UCF Sports\\UCF101\end{tabular} & RGB & Weekly supervised based & SVM & \begin{tabular}{@{}c@{}}98.27\\84.72\end{tabular} \\ \hline
Al et al. \cite{al2021making} & 2021 & \begin{tabular}{@{}c@{}}E-KTH\\E-UCF11\\E-HMDB51\\E-UCF50\\R-UCF11\\R-UCF50\\N-Actions\end{tabular} & RGB & \begin{tabular} {@{}c@{}}  Local and global\\ feature extraction \end{tabular} & QSVM & \begin{tabular}{@{}c@{}}93.14\\94.43\\87.61\\69.45\\82.61\\68.96\\61.94\end{tabular} \\ \hline
Hejazi et al. \cite{hejazi2022handcrafted} & 2022 & \begin{tabular}{@{}c@{}}UCF101\\Kinetics-400\\Kinetics-700\end{tabular} & RGB & Optical flow based & KNN & \begin{tabular}{@{}c@{}}99.21\\98.24\\96.35\end{tabular} \\ \hline
Zhang et al. \cite{zhang2022hybrid} & 2022 & \begin{tabular}{@{}c@{}}UCF 11\\UCF 50\\UCF 101\\JHMDB51\\UT-Interaction\end{tabular} & RGB & FV+BoTF & SVM & \begin{tabular}{@{}c@{}}99.21\\92.5\\95.1\\70.8\\91.50\end{tabular} \\ \hline
Fatima et al. \cite{fatima2023novel} & 2023 & UT-Interaction & RGB & SIFT and ORB & Decision Tree & 94.6 \\ \hline
\end{tabular}

\end{table}

\begin{figure*}[ht]
\begin{adjustwidth}{-2cm}{0cm}
    \centering
    \includegraphics[scale=0.70]{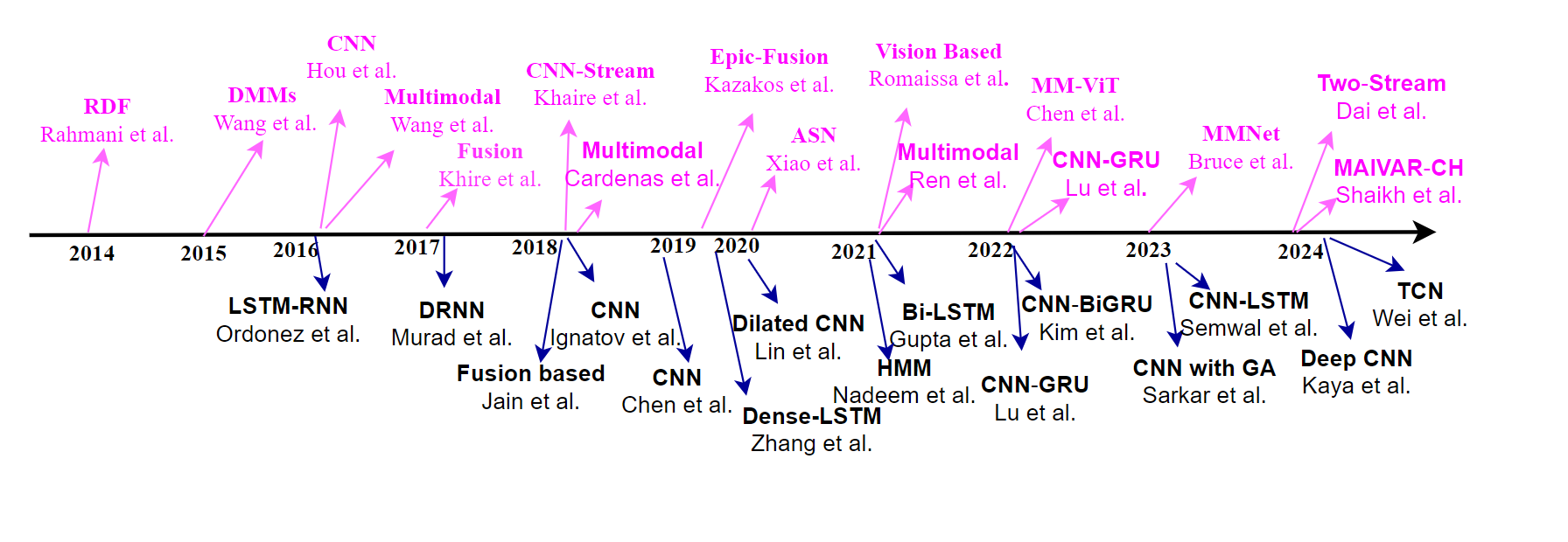}
    \caption{Milestone approaches for HAR. The pink font is the multimodality-based method and the black font is the sensor-based method. }
    \label{fig_milestone_sensor_multimodal}
   \end{adjustwidth}
\end{figure*}
\subsection{End-to-End Deep Learning Approach}
\label{section_}
The holistic, local, and global features reported promising results in the HAR task, but these handcrafted features need much specific knowledge to define relevant parameters. Additionally, they do not generalize the sizeable data set well. In recent years, significant focus has been on utilizing DL in computer vision. Numerous approaches have been used deep neural network-based to recognize human activity \cite{kar2017adascan,varol2017long, feichtenhofer2017spatiotemporal} \cite{simonyan2014two,liu2016spatio,wang2015action,yan2018spatial,tran2015learning,zhang2018real,miah2022bensignnet_miah,miah2024hand_multiculture_miah}.
Figure \ref{fig_milestone_RGB_skeleton} demonstrates the year-wise end-to-end deep learning method developed by various researchers for the RGB-based HAR systems. 
Recently, researchers have utilized different ideas for spatiotemporal feature extraction, divided into three categories: two-stream networks, multi-stream networks, 3D CNN, and Hybrid Networks.   

\begin{figure*}[ht]
\begin{adjustwidth}{-2cm}{0cm}  
    \centering
    \includegraphics[scale=0.30]{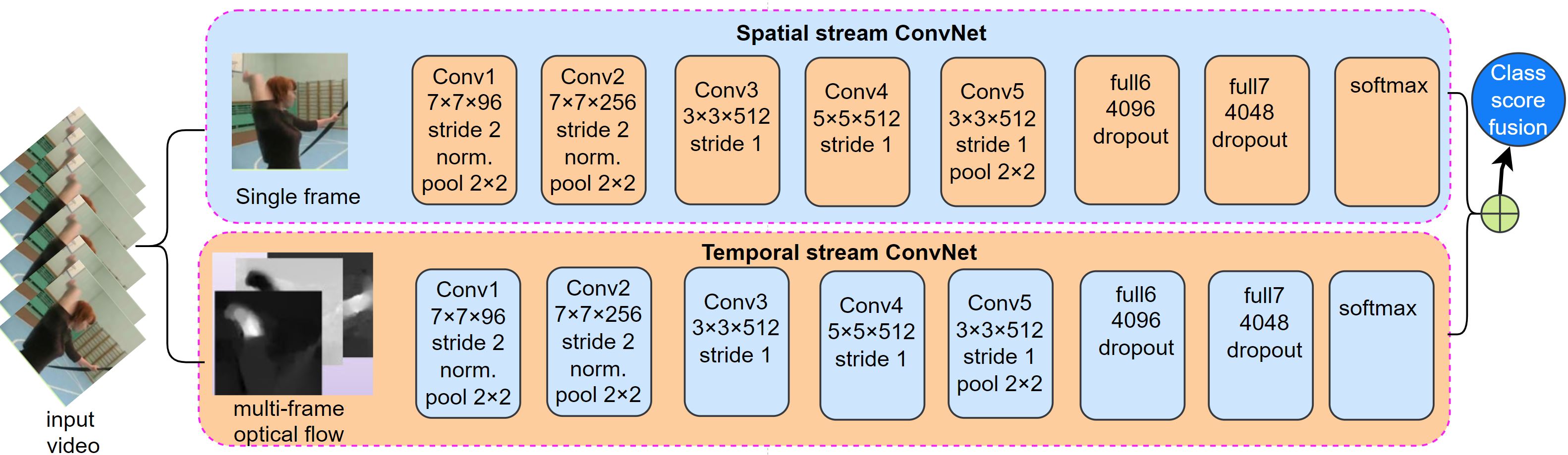}
    \caption{RGB Based Two-stream architecture HAR.}
    \label{fig_RGB_two_stream}
    \end{adjustwidth}
\end{figure*}

\subsubsection{Two Stream Based Network} 
The motion of an object can be represented based on the optical flow \cite{horn1981determining}. 
Simonyan et al. proposed a two-stream convolutional network to recognize human activity \cite{simonyan2014two} as depicted in Figure \ref{fig_RGB_two_stream}.
In a convolutional network with two streams, the optical flow information is computed from the sequence of images. Two separate CNNs process image and optical flow sequences as inputs during model training. Fusion of these inputs occurs at the final classification layer. The two-stream network handles a single-frame image and a stack of optical flow frames using 2D convolution. In contrast, a 3D convolutional network treats the video as a space-time structure and employs 3D convolution to capture human action features.\\
Numerous research endeavors have been conducted to enhance the efficacy of these two network architectures. Noteworthy advancements in the two-stream CNNs have been made by Zhang et al.\cite{zhang2018real}, who substituted the optical flow sequence with the motion vector in the video stream. This substitution resulted in improved calculation speed and facilitated real-time implementation of the aforementioned HAR technique. The process of merging spatial and temporal information has been modified by Feichtenhofer et al. \cite{feichtenhofer2016convolutional}, shifting it from the initial final classification layer to an intermediate position within the network. As a result, the accuracy of action recognition has been further enhanced. The input structure of the convolutional network, as well as the strategy for training, have been extensively examined by Wang et al.\cite{wang2016temporal}. Moreover, an additional enhancement to the performance of the two-stream convolutional network was introduced through the proposal of a temporal segment network (TSN). Moreover, the recognition results of TSN were further improved by the contributions of both Lan et al. \cite{lan2017deep} and Zhou et al. \cite{zhou2018temporal}. 
Depending on the architecture of the deep learning network, notable works typically focus on methods using two-stream CNNs \cite{simonyan2014two}. Transfer learning with RGB data enhances action recognition by leveraging pre-trained models' knowledge. Pham et al. \cite{pham2020unified} present a DL-based framework where poses extracted from RGB video sequences are converted into image-based representations and inputted into a deep CNN, utilizing attention mechanisms to highlight critical features.
\subsubsection{Multi Stream Based Network}
RGB data paired with  CNNs offers powerful action recognition capabilities. Liu et al. \cite{liu2017enhanced} leverage a multi-stream convolutional network to enhance recognition performance by incorporating manually crafted skeleton joint information with CNN-derived features. Shi et al. \cite{shi2017sequential} employ transfer learning techniques in a three-stream network, incorporating dense trajectories to characterize long-term motion effectively.
Attention mechanisms with RGB data focus on relevant regions for better action recognition. 

\begin{table}[!htp]
\centering
\setlength{\tabcolsep}{2pt}
\caption{RGB and deep learning-based existing techniques for action recognition.}
\begin{tabular}{|c|c|c|c|c|c|c|}
\hline
\textbf{Author} & \textbf{Year} & \textbf{Dataset Name} & \textbf{Modality} & \textbf{Method} & \textbf{Classifier} & \textbf{Accuracy [\%]} \\ \hline
Ji et al. \cite{ji20123d} & 2012 & KTH & RGB & 3DCNN & & 90.2 \\ \hline
Wang et al. \cite{wang2015action} & 2015 & \begin{tabular}{@{}c@{}}UCF101\\HMDB51\end{tabular} & RGB & \begin{tabular} {@{}c@{}} 2-stream \\Convolution Network  \end{tabular} & SoftMax & \begin{tabular}{@{}c@{}}91.5\\65.9\end{tabular} \\ \hline
Sharma et al. \cite{sharma2015action} & 2015 & \begin{tabular}{@{}c@{}}UCF11\\HMDB51\\Hollywood2\end{tabular} & RGB & Stacked LSTM & SoftMax & \begin{tabular}{@{}c@{}}84.96\\41.31\\43.91\end{tabular} \\ \hline
Ijjina et al. \cite{ijjina2016human} & 2016 & UCF50 & RGB & CNN-Genetic Algorithm & CNN & 99.98 \\ \hline
Feichtenhofer et al. \cite{feichtenhofer2016convolutional} & 2016 & \begin{tabular}{@{}c@{}}UCF101\\HMDB51\end{tabular} & RGB & CNN Two-Stream & SoftMax & \begin{tabular}{@{}c@{}}92.5\\65.4\end{tabular} \\ \hline
Wang et al. \cite{wang2016temporal} & 2016 & \begin{tabular}{@{}c@{}}HMDB51\\UCF101\end{tabular} & RGB & TSN & SoftMax & \begin{tabular}{@{}c@{}}69.4\\94.2\end{tabular} \\ \hline
Akilan et al. \cite{akilan2017late} & 2017 & \begin{tabular}{@{}c@{}}CIFAR100\\Caltech101\\CIFAR10\end{tabular} & RGB & ConvNets & SoftMax & \begin{tabular}{@{}c@{}}75.87\\95.54\\91.83\end{tabular} \\ \hline
Shi et al. \cite{shi2017sequential} & 2017 & \begin{tabular}{@{}c@{}}KTH\\UCF101\\HMDB51\end{tabular} & RGB & 3-stream CNN & SoftMax & \begin{tabular}{@{}c@{}}96.8\\94.33\\92.2\end{tabular} \\ \hline
Ahsan et al. \cite{ahsan2018discrimnet} & 2018 & \begin{tabular}{@{}c@{}}UCF101\\HMDB51\end{tabular} & RGB & GAN & SoftMax & \begin{tabular}{@{}c@{}}47.2\\41.40\end{tabular} \\ \hline
Tu et al. \cite{tu2018multi} & 2018 & \begin{tabular}{@{}c@{}}JHMDB\\HMDB51\\UCF Sports\\UCF101\end{tabular} & RGB & Multi-Stream CNN & SoftMax & \begin{tabular}{@{}c@{}}71.17\\69.8\\58.12\\94.5\end{tabular} \\ \hline
Zhou et al. \cite{zhou2018mict} & 2018 & \begin{tabular}{@{}c@{}}HMDB51\\UCF101\end{tabular} & RGB & TMiCT-Net & CNN & \begin{tabular}{@{}c@{}}70.5\\94.7\end{tabular} \\ \hline
Jian et al. \cite{jian2019deep} & 2019 & Sport video & RGB & FCN & SoftMax & 97.40 \\ \hline
Ullah et al. \cite{ullah2019action} & 2019 & \begin{tabular}{@{}c@{}}UCF50\\UCF101\\YouTube action\\HMDB51\end{tabular} & RGB & Deep autoencoder & SVM & \begin{tabular}{@{}c@{}}96.4\\94.33\\96.21\\70.33\end{tabular} \\ \hline
Gowda et al. \cite{gowda2012smart} & 2020 & \begin{tabular}{@{}c@{}}UCF101\\HMDB51\\FCVID\\ActivityNet\end{tabular} & RGB & SMART & SoftMax & \begin{tabular}{@{}c@{}}98.6\\84.3\\82.1\\84.4\end{tabular} \\ \hline
Khan et al. \cite{khan2020human} & 2020 & \begin{tabular}{@{}c@{}}HMDB51\\UCF Sports\\YouTube\\IXMAS\\KTH\end{tabular} & RGB & VGG19 CNN & Naive Bayes & \begin{tabular}{@{}c@{}}93.7\\98.0\\94.4\\99.4\\95.2\\97.0\end{tabular} \\ \hline
Ullah et al. \cite{ullah2021efficient} & 2021 & \begin{tabular}{@{}c@{}}HMDB51\\UCF101\\UCF50\\Hollywood2\\YouTube Actions\end{tabular} & RGB & DS-GRU & SoftMax & \begin{tabular}{@{}c@{}}72.3\\95.5\\95.2\\71.3\\97.17\end{tabular} \\ \hline
Wang et al. \cite{wang2021tdn} & 2021 & \begin{tabular}{@{}c@{}}SomethingV1\\SomethingV2\\Kinetics-400\end{tabular} & RGB & Temporal Difference Networks & TDN & \begin{tabular}{@{}c@{}}\\84.1\\91.6\\94.4\end{tabular} \\ \hline
Wang et al. \cite{wang2022hybrid} & 2022 & UCF101 & RGB & HyRSM & - & 93.0 \\ \hline
Wensel et al. \cite{wensel2023vit} & 2023 & \begin{tabular}{@{}c@{}}YouTube Action\\HMDB51\\UCF50\\UCF101\end{tabular} & RGB & ViT-ReT & SoftMax & \begin{tabular}{@{}c@{}}92.4\\78.4\\97.1\\94.7\end{tabular} \\ \hline
Hassan et al. \cite{hassan2024deep} & 2024 & \begin{tabular}{@{}c@{}}UCF11\\UCF Sports\\JHMDB\end{tabular} & RGB & Deep Bi-LSTM & SoftMax & \begin{tabular}{@{}c@{}}99.2\\93.3\\76.3\end{tabular} \\ \hline
\bottomrule
\end{tabular}
\label{Tab:RGB_based}
\end{table}

\subsubsection{3D CNN and Hybrid Networks}
\begin{figure}[ht]
    \centering
    \includegraphics[scale=0.45]{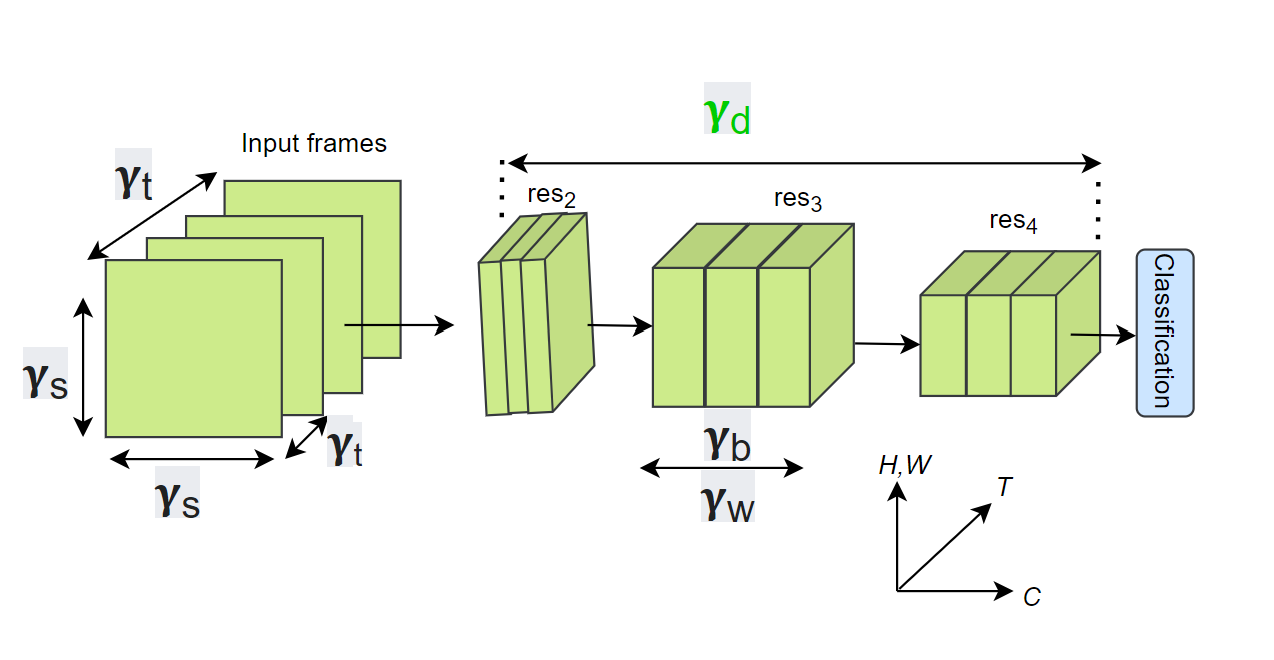}
    \caption{The X3D model framework.}
    \label{fig1_frame X3D}
\end{figure}

Traditional two-stream techniques often separate spatial and temporal information, which can render them less suitable for real-time deployment. However, subsequent research introduced 3D convolutional approaches that directly extract information across all three dimensions. These 3D approaches aim to address the limitations of the earlier two-stream networks.
Ji et al. \cite{ji20123d} utilized the  3D CNN model for the action recognition task. This model extracts features from both the spatial and the temporal dimensions.
Tran et al. \cite{tran2015learning} used C3D  to extract spatiotemporal features for a large dataset to train the model, which is the extension of the 3DCNN model \cite{ji20123d}. 
Carreira et al. \cite{carreira2017quo} developed I3D, extending the network to extract spatiotemporal features along with temporal dimension. They proposed image classification models to create 3D CNNs by transferring weights from 2D models pre-trained on ImageNet to align with the weights in the 3D model. P3D \cite{qiu2017learning} and R(2+1)D \cite{tran2018closer} streamlined 3D network training using factorization, combining 2D spatial convolutions (1×3) with 1D temporal convolutions (3×1×1) instead of traditional 3D convolutions (3×3). For improved motion analysis, trajectory convolution \cite{zhao2018trajectory} employed deformable convolutions in the temporal domain. Other approaches simplify 3D CNNs by integrating 2D and 3D convolutions within single networks to enhance feature maps, exemplified by models like MiCTNet \cite{zhou2018mict}, ARTNet \cite{wang2018appearance}, and S3D \cite{xie2018rethinking}.
To enhance the performances of 3DCNN, CSN \cite{tran2019video}  has demonstrated the effectiveness of decomposing 3D convolution by separating channel interactions from spatiotemporal interactions, leading to state-of-the-art performance improvements. This technique can achieve speeds 2 to 3 times faster than previous methods. Feichtenhofer et al. developed the X3D methods \cite{feichtenhofer2020x3d} as shown in Figure \ref{fig1_frame X3D}. The X3D network included both spatial and temporal dimensions with enhanced spatial, input resolution, and channel dimensions. 
Yang et al. \cite{yang2020temporal} proposed that morphologically similar actions like walking, jogging, and running require discrimination assisted by visual speed. They proposed a Temporal Pyramid Network (TPN) similar to X3D. This approach enables the extraction of effective features at various temporal rates, reducing computational complexity while enhancing efficiency performances.
Zhang et al. \cite{zhang2020v4d}  proposed a 4D CNN with 4D convolution to capture the evolution of distant spatiotemporal representations.\\ 
Similarly, numerous researchers have made efforts to expand various 2D CNNs to 3D spatiotemporal structures to acquire knowledge about and identify human action features, drawing inspiration from the concept of 3D (Three-dimensional) ConvNet. Carreira et al. \cite{carreira2017quo} expanded the network architecture of inception-V1 to incorporate  3D and introduced the two-stream inflated 3D ConvNet for HAR.
Qin et al. \cite{qin2017feature} propose a fusion scheme combining classical descriptors with 3D CNN-learned features, achieving robustness against geometric and optical deformations. Diba et al. \cite{diba2017temporal} extended DenseNet and introduced a temporal 3D ConvNet for HAR. Zhu et al. \cite{zhu2018end} expanded pooling operations across spatial and temporal dimensions, transforming the two-stream convolution network into a three-dimensional structure. Carreira et al. \cite{carreira2017quo} conducted a comparison of five architectures: LSTM with CNN, 3D ConvNet, two-stream network, two-stream inflated 3D ConvNet, and 3D-fused two-stream network. In essence, 3D CNNs establish relationships between temporal and spatial features in various ways, complementing rather than replacing two-stream networks.
Hassan et al. \cite{hassan2024deep} a deep bidirectional LSTM model, which effectively integrates the advantages of temporal effective features extraction through bi-LSTM and spatial feature extraction via CNN. The LSTM  architecture is not feasible to support parallel computing, which can limit its efficiency. To overcome this problem, the transformer architecture \cite{vaswani2017attention} has become popular in DL to address this limitation. Girdhar et al. \cite{girdhar2019video} used the transformer-based architecture to add context features and developed an attention mechanism to improve performance.
 
\subsubsection{Recurrent Neural Network Based Approach}
Unlike two-stream and 3D ConvNet, which use various convolutional temporal feature pooling architectures to model action, LSTM-based methods well perform view a video as a sequential arrangement of frames. The representation of HAR can subsequently be depicted through the alterations in features observed in each frame. Donahue et al. explored LSTM and developed LRCN \cite{donahue2015long} to model CNN-generated spatial features across temporal sequences.
Another significant HAR technique involves the use of LSTM with CNNs \cite{carreira2017quo,ng2015beyond}. Ng et al. \cite{ng2015beyond} introduced a recurrent neural network (RNN) model to identify and classify the action, which performs a connection between the LSTM cell and the output of the underlying CNN. Furthermore, Qiu et al.\cite{qiu2017learning}  proposed a novel architectural design termed Pseudo-3D ResNet (P3D ResNet), wherein each block is assembled in a distinct ResNet configuration. Donahue et al. \cite{donahue2015long} proposed a method of using long-term RNNs to map video frames of varying lengths to outputs of varying lengths, such as action descriptive text, rather than simply assigning them to a specific action category. Song et al. \cite{song2018spatio} introduced a model using RNNs with LSTM that employed multiple attention levels to discern key joints in the skeleton across each input frame.


\section{Skeleton Data Modality Based Action Recognition Method}\label{section_3}
The main challenges of the RGB-based data modality-based HAR system are redundant background and computational complexity issues, and the Skeleton-based data modality helps us overcome these challenges. In addition, coupled with joint coordinate estimation algorithms such as OpenPose and SDK \cite{han2013enhanced} has improved the performance of accuracy and reliability of the skeleton data. Skeleton data obtained from the joint position offers several benefits over the RGB data, such as illumination variations, viewing angles, and background occlusions, making it less susceptible to noise interference. The research prefers to perform HAR by using the skeleton data because it provides more focused information and reduces redundancy. Based on the feature extraction methods for HAR, the skeleton data can be divided into DL-based methods, relying on learned features, and ML-based methods, which use handcrafted features. In addition, the skeleton data depends on the precise joint position and pose estimation techniques.

\begin{figure}[ht]
    \centering
    \includegraphics[scale=0.45]{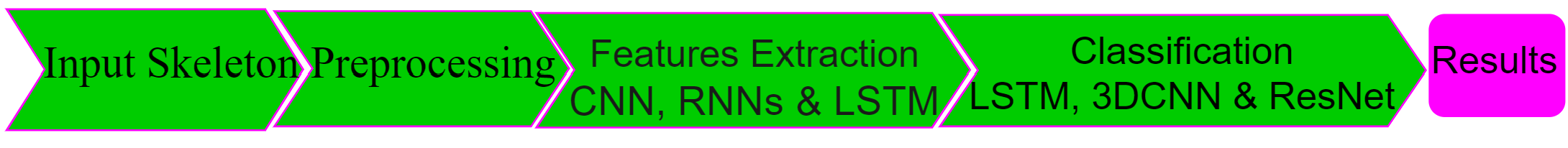}
    \caption{Skeleton-based action recognition.}
    \label{fig1_Skeleton}
\end{figure}

 Figure \ref{fig1_Skeleton} shows the framework of skeleton-based approaches. Table \ref{tab:sekelton_comparative_analysis} describes the key information about the skeleton-based data modality on the existing model, including datasets, classification methods, years, and performance accuracy. We describe the well-known pose estimation algorithms in the following section.

\subsection{Skeleton Based HAR Dataset}
We provided the most popular benchmark HAR datasets, which come from the skeleton, which is demonstrated in Table \ref{tab1:rgb_skeleton_dataset}. The dataset table demonstrated the details of the datasets, including modalities, creation year, number of classes, number of subjects who participated in recording the dataset, number of samples, and latest performance accuracy of the dataset with citation. 
The Skeleton dataset includes a variety of notable benchmarks essential for Human Activity Recognition (HAR). The UPCV dataset from 2014 features 10 classes, 20 subjects, and 400 samples, achieving an outstanding accuracy of 99.2\% \cite{theodorakopoulos2014pose} \cite{zhou2022high}. The NTU RGB+D dataset, introduced in 2016 and expanded in 2019, is one of the most comprehensive, with 60 and 120 classes, 40 and 106 subjects, and 56,880 and 114,480 samples, respectively, both versions recording an accuracy of 97.4\% \cite{shahroudy2016ntu} \cite{liu2019ntu} \cite{cheng2024dense}. The MSRDailyActivity3D dataset from 2012 includes 16 classes, 10 subjects, and 320 samples, with an accuracy of 97.5\% \cite{wang2012mining} \cite{shahroudy2017deep}. The PKU-MMD dataset from 2017 contains 51 classes, 66 subjects, and 10,076 samples, with a notable accuracy of 94.4\% \cite{liu2017pku} \cite{li2019making}. The Multi-View TJU dataset from 2014 offers 20 classes, 22 subjects, and 7,040 samples. These datasets are crucial for training and testing HAR models, offering diverse activities and scenarios to enhance model robustness and accuracy.
\subsection{Pose Estimation}
\begin{figure}[t] 
    \centering
    \includegraphics[scale=0.20]{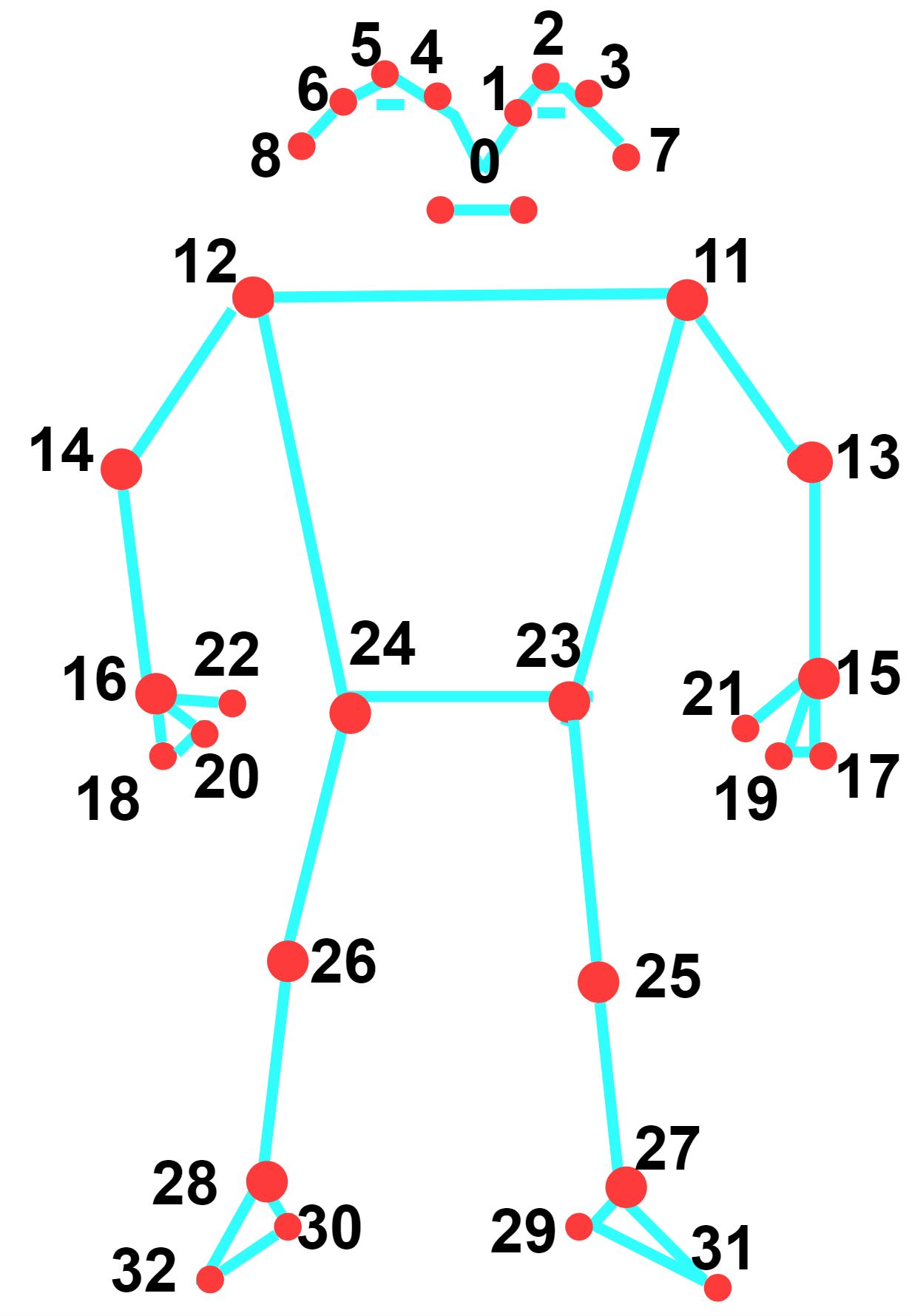}
    \caption{Landmarks position.}
    \label{fig_landmark_position} 
\end{figure}

We can extract human joint skeleton points from the RGB video using media pipe, openpose, AlphaPose ~\cite{fang2017rmpe,xiu2018pose}, MMPose, etc.  Using a media pipe, figure \ref{fig_landmark_position} demonstrated the 33 joint skeleton points from the whole body. Human limb trunk reconstruction included estimating human pose by detecting joint positions in the skeleton and establishing their connections. Traditional methods, relying on manual feature labeling and regression for joint coordinate retrieval, suffer from low accuracy. DL-based methods, including 2D and 3D pose estimation, have become pivotal in this research domain.

\subsubsection{2D Human Pose Estimation Based Methods}
The objective of 2D human pose estimation is to identify significant body parts in an image and connect them sequentially to form a human skeleton graph. Research commonly addresses the classification of single and multiple human subjects. In single-person pose estimation, the goal is to detect a solitary individual in an image. This involves first recognizing all joints of the person's body and subsequently generating a bounding box around them. Two main categories of models exist for single-person pose estimation. The first utilizes a direct regression approach, where key points are directly predicted from extracted features. In 2D pose estimation, one can employ deformable part models to recognize the object by matching a set of templates. Nevertheless, these deformable part models exhibit limited expressiveness and fail to consider the global context. Yan et al. \cite{yang2012articulated} proposed a pose-based and performed two main methods: detection-based and regression-based approaches. Detection-based methods utilize powerful part detectors based on CNNs, which can be integrated using graphical models as described by Yuille et al. \cite{chen2014articulated}. For solving the detection problem, pose estimation can be represented as a heat map where each pixel indicates the detection confidence of a joint, as outlined by Bulat et al. \cite{bulat2016human}. However, detection approaches do not directly provide joint coordinates. A post-processing step is applied to recover poses where (x, y) coordinates are obtained by utilizing the max function. Toshev et al.\cite{toshev2014deeppose} proposed a cascade of regressor methods to estimate poses, they employ the regression-based approach with a nonlinear function that maps the joint coordinates and refines pose estimates. Carreira et al.\cite{carreira2016human} propose the Iterative Error Feedback (IEF) approach, where iterative prediction is performed to correct the current estimates. Instead of predicting outputs in a single step, a self-correcting model is employed, which modifies an initial solution by incorporating error predictions, also called IEF. However, the sub-optimal nature of the regression function leads to lower performance than detection-based techniques.

\subsubsection{3D Human Pose Estimation Based Methods}
Conversely, when presented with an image containing an individual, the objective of 3D pose estimation is to generate a 3D pose that accurately aligns with the spatial location of the person depicted. The accurate reconstruction of 3D poses from real-life images holds significant potential in various fields of HAR such as entertainment and human-computer interaction, particularly indoors and outdoors. Earlier approaches relied on feature engineering techniques, whereas the most advanced techniques are based on deep neural networks, as proposed by Zhou et al. \cite{zhou2018monocap} 3D pose estimation is acknowledged to be more complex than its 2D handle due to its management of a larger 3D pose space and an increased number of ambiguities. Nunes et al. \cite{nunes2017human} presented skeleton extraction through depth images, wherein skeleton joints are inferred frame by frame. A manually selected set of 15 skeleton joints, as determined by Gan et al. \cite{gan2013human}, they used to form an APJ3D representation, which is based on relative positions and local spherical angles. These 15 joints, which have been deliberately selected, play a crucial role in the development of a concise representation of human posture. Spatial features are encoded using diverse metrics, including joint distances, orientations, vectors, distances between joints and lines, and angles between lines.  These measures collectively contribute to a comprehensive texture feature set, as suggested by Chen et al. \cite{chen2015reduced}. Additionally, a CNN-based network is trained to recognize corresponding actions.

\begin{table}[]
\centering
\setlength{\tabcolsep}{2pt}
\caption{Skeleton and deep learning based on existing approach for action recognition
\label{tab:sekelton_comparative_analysis}}
\begin{tabular}{|c|c|c|c|c|c|c|}
\hline
\textbf{Author} & \textbf{Year} & \textbf{Dataset Name} & \textbf{Modality} & \textbf{Method} & \textbf{Classifier} & \textbf{Accuracy [\%]} \\ \hline
Veeriah et al. \cite{veeriah2015differential} & 2015 & \begin{tabular}{@{}c@{}}MSRAction3D (CV)\\ KTH-1 (CV)\\ KTH-2 (CV)\end{tabular} & Skeleton & Differential RNN &SoftMax & \begin{tabular}{@{}c@{}}92.03 \\ 93.96, 92.12\end{tabular} \\ \hline
Xu et al. \cite{xu2016human} & 2016 & \begin{tabular}{@{}c@{}}MSRAction3D\\ UTKinect\\ Florence3D action\end{tabular} & Skeleton & SVM with PSO &SVM & \begin{tabular}{@{}c@{}}93.75 \\ 97.45, 91.20\end{tabular} \\ \hline
Zhu et al. \cite{zhu2016co} & 2016 & \begin{tabular}{@{}c@{}}SBU Kinect\\ HDM05, CMU\end{tabular} & Skeleton & Stacked LSTM &SoftMAx & \begin{tabular}{@{}c@{}}90.41 \\ 97.25, 81.04\end{tabular} \\ \hline
Li et al. \cite{li2017joint} & 2017 & \begin{tabular}{@{}c@{}}UTD-MHAD\\ NTU-RGBD (CV)\\ NTU-RGBD (CS)\end{tabular} & Skeleton & CNN &Maximum Score& \begin{tabular}{@{}c@{}}88.10 \\ 82.3 \\ 76.2\end{tabular} \\ \hline
Soo et al. \cite{soo2017interpretable} & 2017 & \begin{tabular}{@{}c@{}}NTU-RGBD (CV)\\ NTU-RGBD (CS)\end{tabular} & Skeleton & Temporal CNN &SoftMax & \begin{tabular}{@{}c@{}}83.1 \\ 74.3\end{tabular} \\ \hline
Liu et al. \cite{liu2017enhanced} & 2017 & \begin{tabular}{@{}c@{}}NTU-RGBD (CS)\\ NTU-RGBD (CV)\\ MSRC-12 (CS)\\ Northwestern-UCLA\end{tabular} & Skeleton & Multi-stream CNN &SoftMax& \begin{tabular}{@{}c@{}}80.03, 87.21 \\ 96.62, 92.61\end{tabular} \\ \hline
Das et al. \cite{das2018deep} & 2018 & \begin{tabular}{@{}c@{}}MSRDailyActivity3D\\ NTU-RGBD (CS)\\ CAD-60\end{tabular} & Skeleton & Stacked LSTM &SoftMax & \begin{tabular}{@{}c@{}}91.56 \\ 64.49, 67.64\end{tabular} \\ \hline
Si et al. \cite{si2019attention} & 2019 & \begin{tabular}{@{}c@{}}NTU-RGBD (CS)\\ NTU-RGBD (CV)\\ UCLA\end{tabular} & Skeleton & AGCN-LSTM &Sigmoid& \begin{tabular}{@{}c@{}}89.2, 95.0 \\ 93.3\end{tabular} \\ \hline
Shi et al. \cite{shi2019two} & 2019 & \begin{tabular}{@{}c@{}}NTU-RGBD (CS)\\ NTU-RGBD (CV)\\ Kinetics\end{tabular} & Skeleton & AGCN &SoftMax & \begin{tabular}{@{}c@{}}88.5 \\ 95.1 \\ 58.7\end{tabular} \\ \hline
Trelinski et al. \cite{trelinski2019ensemble} & 2019 & \begin{tabular}{@{}c@{}}UTD-MHAD\\ MSR-Action3D\end{tabular} & Skeleton & CNN-based &SoftMax & \begin{tabular}{@{}c@{}}95.8, 77.44 \\ 80.36\end{tabular} \\ \hline
Li et al. \cite{li2019actional} & 2019 & \begin{tabular}{@{}c@{}}NTU-RGBD (CS)\\ Kinetics (CV)\end{tabular} & Skeleton &  \begin{tabular}{@{}c@{}}Actional graph \\ based CNN\end{tabular}  &SoftMax & \begin{tabular}{@{}c@{}}86.8 \\ 56.5\end{tabular} \\ \hline
Huynh et al. \cite{huynh2019encoding} & 2019 & \begin{tabular}{@{}c@{}}MSRAction3D\\ UTKinect-3D\\ SBU-Kinect Interaction\end{tabular} & Skeleton & ConvNets &SoftMax & \begin{tabular}{@{}c@{}}97.9 \\ 98.5, 96.2\end{tabular} \\ \hline
Huynh et al. \cite{huynh2020image} & 2020 & \begin{tabular}{@{}c@{}}NTU-RGB+D\\ UTKinect-Action3D\end{tabular} & Skeleton & PoT2I with CNN & SoftMax& \begin{tabular}{@{}c@{}}83.85,98.5\end{tabular} \\ \hline
Naveenkumar et al. \cite{naveenkumar2020deep} & 2020 & \begin{tabular}{@{}c@{}}UTKinect-Action3D\\ NTU-RGB+D\end{tabular} & Skeleton & Deep ensemble &SoftMax & \begin{tabular}{@{}c@{}}98.9, 84.2\end{tabular} \\ \hline
Plizzari et al. \cite{plizzari2021skeleton} & 2021 & \begin{tabular}{@{}c@{}}NTU-RGBD 60\\ NTU-RGBD 120\\ Kinetics Skeleton-400\end{tabular} & Skeleton & ST-GCN &SoftMax& \begin{tabular}{@{}c@{}}96.3, 87.1\\60.5\end{tabular} \\ \hline
Snoun et al. \cite{Snoun2021HAR} & 2021 & \begin{tabular}{@{}c@{}}RGBD-HuDact,  KTH\end{tabular} & Skeleton & VGG16 & SoftMax& \begin{tabular}{@{}c@{}}95.7, 93.5\end{tabular} \\ \hline
Duan et al. \cite{duan2022pyskl} & 2022 & \begin{tabular}{@{}c@{}}NTU-RGBD\\ UCF101\end{tabular} & Skeleton & PYSKL &- & \begin{tabular}{@{}c@{}}97.4, 86.9\end{tabular} \\ \hline
Song et al. \cite{song2022constructing} & 2022 & \begin{tabular}{@{}c@{}}NTU-RGBD\end{tabular} & Skeleton & GCN &SoftMax& \begin{tabular}{@{}c@{}}96.1\end{tabular} \\ \hline
Zhu et al. \cite{Zhu2023} & 2023 & \begin{tabular}{@{}c@{}}UESTC\\ NTU-60 (CS)\end{tabular} & Skeleton & RSA-Net &SoftMax & \begin{tabular}{@{}c@{}}93.9, 91.8\end{tabular} \\ \hline
Zhang et al. \cite{zhang2023fast} & 2023 & \begin{tabular}{@{}c@{}} NTU-RGBD\\ Kinetics-Skeleton\end{tabular} & Skeleton & Multilayer LSTM &SoftMax & \begin{tabular}{@{}c@{}}83.3\\ 27.8(Top-1)\\ 50.2( Top-5)\end{tabular} \\ \hline
Liu et al. \cite{liu2023skeleton} & 2023 & \begin{tabular}[c]{@{}c@{}}NTU-RGBD 60\\ (CV)NTU-RGBD 120 (CS)\end{tabular} & Skeleton & LKJ-GSN &SoftMax & \begin{tabular} {@{}c@{}} 96.1\\86.3\end{tabular} \\ \hline
Liang et al. \cite{Liang2024} & 2024 & \begin{tabular}[c]{@{}c@{}}NTU-RGBD (CV)\\ NTU-RGBD 120 (CS)\\ FineGYM\end{tabular} & Skeleton & MTCF & SoftMax& \begin{tabular} {@{}c@{}} 96.9, 86.6\\94.1\end{tabular} \\ \hline
\end{tabular}
\end{table}

\subsection{Handcrafted Feature and ML Based Classification Approach} 
Researchers determine handcrafted features using statistical features extracted from action data. These features describe the dynamics or statistical properties of the action analyzed. Yang et al. \cite{yang2014super} proposed a method to extract the super vector features to determine the action based on the depth information. Shao et al. \cite{shao2012human} combine shape and motion information for HAR through temporal segmentation, utilizing MHI and Predicted Gradients (PCOG) as feature descriptors. Yang et al. \cite{yang2012recognizing} introduced the depth motion map (DMM) technique, which allows for the projection and compression of the spatiotemporal depth structure from different viewpoints, including the side, front, and upper views. This process results in the formation of three distinct motion history maps. To represent these motion history maps, the authors employed the HOG feature. Instead of using HOG, Chen et al. \cite{chen2015action} employed local binary pattern features to describe human activities based on Dynamic Motion Models (DMMs). Additionally, Chen et al. \cite{chen2015triviews} introduced a spatiotemporal depth layout across frontal, lateral, and upper orientations. Departing from depth compression methods, they extracted motion trajectory shapes and boundary histogram features from spatiotemporal interest points, leveraging dense sampling and joint points in each perspective to depict actions. Moreover, Miao et al. \cite{miao2016efficient} applied the discrete cosine variation technique for effective compression of depth maps. Simultaneously, they generated action features by utilizing transform coefficients. From the available depth data, it is possible to estimate the structure of the human skeleton promptly and precisely.  Shotton et al. \cite{shotton2013real} proposed a method for real-time estimation of body postures from depth images, thereby facilitating rapid segmentation of humans based on depth. Within this context, the problem of detecting joints has been simplified to a per-pixel classification task. Additionally, there is ongoing research in the field of HAR that employs depth data and focuses on methods utilizing the human skeleton. These approaches analyze changes in the joint points of the human body across consecutive video frames to characterize actions, encompassing alterations in both the position and appearance of the joint points. Xia et al. \cite{xia2012view} proposed a three-dimensional joint point histogram as a means to depict the human pose and subsequently formulated the action using a discrete hidden Markov model.  Keceli et al. \cite{keceli2014recognition}, captured depth and human skeleton information via the employment of the Kinect sensor, and subsequently derived human action features by assessing the angle and displacement information about the skeleton joint points. Similarly, Yang et al. \cite{yang2014effective} developed a method based on the EigenJoints, which leverages an accumulative motion energy (AME) function to identify video frames and joint points that offer richer information for action modeling. Pazhoumand et al. \cite{pazhoumand2015joint} utilized the longest common subsequent method to select distinctive features with high discriminatory power from the skeleton's relative motion trajectories, thereby providing a comprehensive description of the corresponding action.\\
Handcrafted features offer high interpretability, simplicity, and straight-forward. However, the handcrafted features-based method requires prior knowledge, which is difficult to generalize. 



\subsection{End to End Deep Learning Based Approach}
Recently, there has been a growing HAR of the advantages of integrating skeleton data with DL-based techniques. The handcrafted features have reduced discriminative capability for  HAR; conversely, to extract features efficiently, the utilization of methods based on DL  necessitates a substantial quantity of training data. Figure \ref{fig_milestone_RGB_skeleton} also demonstrates the year-wise end-to-end deep learning method developed by various researchers for the skeleton-based HAR systems.  As shown, several notable models leveraging recurrent neural networks (RNN), CNN, and graph convolutional networks (GCN) have developed. 
\subsubsection{CNN-Based Methods}
Skeleton data combined with ML methods provides efficient action recognition capabilities. Zhang et al. \cite{zhang2012microsoft} utilize the Kinect sensor to capture skeletal representations, enabling the recognition of actions based on body part movements. Skeleton data paired with CNNs offers robust action recognition. 
As a result, in the work of Wang et al. \cite{wang2015action}, an advantage is found in combining handcrafted and DL-based features through the use of an enhanced trajectory. Additionally, the Trajectory-pooled Deep-Convolutional Descriptor (TpDD), also referred to as Two-stream ConvNets is employed. The construction of an effective descriptor is achieved through the learning of multi-scale convolutional feature maps within a deep architecture. 
Ding et al. \cite{ding2017investigation} developed a CNNs-based model to extract high-level effective semantic features from RGB textured images obtained from using skeletal data. However, these methodologies have a lot of preprocessing steps and a chance to miss some effective information. Caetano et al. suggested SkeleMotion \cite{caetano2019skelemotion}, which offers a novel skeleton image representation as an alternative input for neural networks to address these issues. Researchers have explored solutions to the challenge of long-time dependence, especially considering that CNN did not extract long-distance motion information. To overcome this issue Liu et al. \cite{liu2017two} suggested a Subsequence Attention Network (SSAN) to improve the capture of long-term features. This network, combined with 3DCNN, uses skeleton data to record long-term features more effectively.  

\subsubsection{RNN-LSTM Based Methods}
Approaches relying on Recurrent Neural Networks with LSTM units (RNN-LSTM) \cite{hochreiter1997long,ogiela2012computational}  have garnered considerable popularity as a predominant DL methodology for skeleton-based action recognition. Moreover, these approaches have demonstrated exceptional proficiency in accomplishing video-based action recognition tasks \cite{du2015hierarchical,veeriah2015differential,shahroudy2016ntu,liu2016spatio,zhu2016co,liu2017global}. The spatio-temporal patterns of skeletons exhibit temporal evolutions. Consequently, these patterns can be effectively represented by memory cells within the structure of RNN-LSTM models, as proposed by \cite{hochreiter1997long}. In a similar vein, Du et al. \cite{du2015hierarchical} introduced a hierarchical RNN approach to capture the long-term contextual information of skeletal data. This involved dividing the human skeleton into five distinct parts based on its physical structure. Subsequently, each lower-level part was represented using an RNN, and these representations were then integrated to form the final representation of higher-level parts, which facilitated action classification. 
The problem related to gradient explosion and vanishing gradients occurs if the sequences are too long for actual training. To overcome this issue li et al. \cite{li2018independently} suggested an independent recurrent neural network (IndRNN) to regulate gradient backpropagation over time, allowing the network to capture long-term dependencies.
Shahroudy et al. \cite{shahroudy2016ntu} introduced a model for human action learning using a part-aware LSTM. This model involves splitting the long-term memory of the entire motion into part-based cells and independently learning the long-term context of each body part. The network's output is then formed by combining the independent body part context information. Liu et al. \cite{liu2016spatio} presented a spatio-temporal LSTM network named ST-LSTM, which aims at 3D action recognition from skeletal data. They proposed a technique called skeleton-based tree traversal to feed the structure of the skeletal data into a sequential LSTM network and improved the performance of ST-LSTM by incorporating additional trust gates.
In their recent work, Liu et al. \cite{liu2017global} directed their attention towards the selection of the most informative joints in the skeleton by employing a novel type of LSTM network called Global Context-Aware Attention  (GCA-LSTM) to recognize actions based on 3D skeleton data. Two layers of LSTM were utilized in his study. The initial layer encoded the input sequences and produced a global context memory for these sequences. Simultaneously, the second layer carried out attention mechanisms over the input sequences with the support of the acquired global context memory. The resulting attention representation was subsequently employed to refine the global context. Numerous iterations of attention mechanisms were conducted, and the final global contextual information was employed in the task of action classification. Compared to the methodologies based on hand-crafted designed local features, the RNN-LSTM methodologies and their variations have demonstrated superior performance in recognition of actions.
Nevertheless, these methodologies tend to excessively emphasize the temporal information while neglecting the spatial information of skeletons \cite{du2015hierarchical,veeriah2015differential,shahroudy2016ntu,liu2016spatio,zhu2016co,liu2017global}. RNN-LSTM methodologies continue to face difficulties in dealing with the intricate spatio-temporal variations of skeletal movements due to multiple issues, such as jitters and variability in movement speed. Another drawback of the RNN-LSTM networks \cite{hochreiter1997long,ogiela2012computational} is their sole focus on modelling the overall temporal dynamics of actions, disregarding the detailed temporal dynamics. To address these limitations, in this investigation, a CNN-based methodology can extract discriminative characteristics of actions and model various temporal dynamics of skeleton sequences via the suggested Enhanced-SPMF representation, encompassing short-term, medium-term, and long-term actions.

\subsubsection{GNN or GCN-Based Methods}
Graph convolutional neural networks (GCNNs) are powerful DL-based methods designed to perform non-Euclidean data. Unlike traditional  CNNs and RNNs, which perform well with Euclidean data (such as images, text, and speech), they are unable to perform with non-Euclidean data \cite{miah2023dynamic_miah,miah2024hand_multiculture_miah,miah2024spatial_paa_paa,10360810_ksl2_miah,10510436_anomaly_miah,egawa2023dynamic_fall_miah,miah2024sign_largescale_miah,electronics12132841_miah_multistream_4}. The GCN was first introduced by Gori et al. \cite{gori2005new} in 2005 to handle graph data. GCNNs with skeleton data enable spatial dependencies to be captured for accurate action recognition. The human skeleton data, consisting of joint points and skeletal lines, can be viewed as non-Euclidean graph data. Therefore, GCNs are particularly suited for learning from such data. There are two main branches of GCNs: Spectral GCN and Spatial GCN.

\begin{itemize}
    \item \textbf{Spectral GCNs based methods:}
    Using and leveraging both eigenvalues and eigenvectors of the graph Laplacian matrix (GLM) to convert graph data from the temporal to the spatial domain \cite{li2017situation}, but this model is not computationally efficient. To address this issue, kipf et al. \cite{kipf2016semi} enhanced the spectral GCN approach by allowing the filter operation of only one neighbour node to reduce the computational cost. While spectral GCNs have shown effectiveness in HAR tasks, their computational cost poses challenges when dealing with graphs.
    \item \textbf{Spatial GCN-based methods:} They are more efficient in terms of computational than spectral GCNs. Therefore, spatial GCNs have become the main focus in many GCN-based HAR approaches due to efficiency. Yan et al. \cite{yan2018spatial} developed the concept of ST-GCN, a model specifically designed for spatiotemporal data. 
\end{itemize}

\begin{figure}[ht]
    \centering
    \includegraphics[scale=0.18]{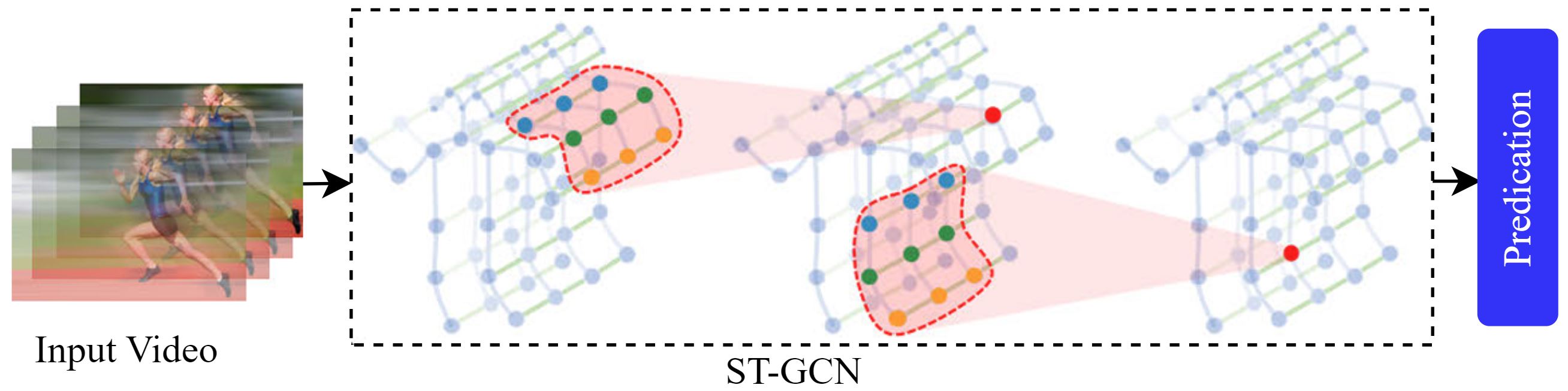}
    \caption{Skeleton-based HAR using ST-GCN.}
    \label{fig_skeleton_stgcn}
\end{figure}

As depicted in Figure \ref{fig_skeleton_stgcn} the  ST-GCN, bodily joints (such as joints in a human skeleton) serve as the vertices in the graph while the edges denote the connection between the bodily bones within the same frame.
Shi et al. \cite{shi2019two} developed two-stream adaptive GCN models to improve the flexibility of graph networks. This model allows for the use of the end-to-end approach to learning the graph's topology within the model. By adopting a data-driven methodology, the 2sAGCN model becomes more adaptable to diverse data samples, increasing flexibility. Additionally, an attention mechanism is included to improve the robustness of the 2sAGCN model. For a further improvement to explore the enhancement of HAR methods, Shiraki et al. \cite{shiraki2020spatial} proposed the spatiotemporal attentional graph (STA)- GCN to determine the challenge varying importance of joints across different human actions. Unlike traditional GCNs, STA-GCN takes into account both the significance and interrelationship of joints within the graph. Researchers have drawn inspiration from STA-GCN to further enhance GCN models \cite{shi2020skeleton,huang2020long}.  For instance, the shift-GCN model introduces the innovative shift-graph method to enhance the flexibility of the spatio-temporal graph's (STG) receptive domain. 
Additionally, the lightweight dot convolution technique is utilized to reduce the number of feature channels and make the model more efficient. Song et al. \cite{song2020stronger} present the residual-based GCN model to improve the performance of the model in terms of accuracy and computational efficiency for HAR. Similarly, Thakkar et al. \cite{thakkar2018part} and Li et al. \cite{li2019spatio} presented methods to divide the human skeleton into separate body parts and they developed the partial-based graph convolutional network (PB-GCN) \cite{thakkar2018part}, which learns four subgraphs of the skeleton data. Li et al. \cite{li2019spatio} developed the spatio-temporal graph routing (STGR) scheme to better determine the connections between joints. These methods help improve the segmentation of body parts for HAR.

\section{Sensor Based HAR}
\label{section_4}
Sensor-based  HAR has gained significant attention due to wearable technology and its applications in various domains. These include health monitoring, industrial safety, sports training, and more \cite{sanhudo2021activity}. Unlike computer vision-based or WIFI-based HAR, wearable sensors offer advantages such as privacy, user acceptance, and independence from environmental factors \cite{chen2021deep}. Challenges in sensor-based HAR include diverse data collection, handling missing values, and complex activity recognition. Wearable devices use sensors like accelerometers and gyroscopes to identify human activities, but feature extraction and model training remain challenging. The challenges with machine learning approaches rely on manual feature extraction \cite{huan2021human} while the DL approaches now enable automatic feature extraction from raw sensor data, leading to superior results \cite{chen2021deep}. Overall, sensor-based HAR holds promise for improving healthcare and safety \cite{nafea2021sensor,miah2017motor,miah2022movie_motor_miah,kabir2023investigating_motor_miah}.\\
Table \ref{tab:sensor_modalitiy_performance} summarizes various existing works based on sensor modality for HAR using traditional  ML and DL techniques, including the author name, year, datasets, modality sensor names, methods, classifier, and performance accuracy. As mentioned in Table, researchers have enhanced HAR classification performance by improving ML feature engineering, and some researchers have developed advanced DL models like CNN and LSTM for automatic feature extraction. Most studies utilized datasets from multiple sensor types placed at different body positions. Additionally,  we summarize several publically available datasets in Table \ref{tab:sensormodilities_dataset}, including year, sensor modalities, number of sensors, number of participants, number of activities, activity categories, and latest performance accuracy.

\begin{table}[]
\centering
\setlength{\tabcolsep}{2pt}
\caption{Databases for Sensor Modality}
\label{tab:sensormodilities_dataset}
\begin{tabular}{|l|l|l|l|l|l|l|l|}
\hline
\begin{tabular}[c]{@{}l@{}}Dataset \\ Names\end{tabular} & 
\begin{tabular}[c]{@{}l@{}}Year \end{tabular} & 
\begin{tabular}[c]{@{}l@{}}Sensor \\ modalities\end{tabular} & 
\begin{tabular}[c]{@{}l@{}}No. of \\ sensors\end{tabular} & 
\begin{tabular}[c]{@{}l@{}}No. of \\ people\end{tabular} & 
\begin{tabular}[c]{@{}l@{}}No. of \\ Activities\end{tabular} & 
\begin{tabular}[c]{@{}l@{}}Activity \\ Categories\end{tabular} &
\begin{tabular}[c]{@{}l@{}}Latest \\ Performances\end{tabular} \\ \hline
HHAR\cite{stisen2015smart} &2015 &
Accelerometer, Gyroscope & 
36 & 
9 & 
6 & 
\begin{tabular}[c]{@{}l@{}}Daily living activity, \\ Sports fitness activity\end{tabular}&\begin{tabular}[c]{@{}l@{}} 99.99\% 
\cite{abbas2024active}\end{tabular} \\ \hline

MHEALTH\cite{banos2014villalonga} &2014 &
\begin{tabular}[c]{@{}l@{}}Accelerometer, Gyroscope, \\ Magnetometer,\\ Electrocardiogram\end{tabular} & 
3 & 
10 & 
12 & 
\begin{tabular}[c]{@{}l@{}}Atomic activity, \\ Daily living activity, \\ Sports fitness activity\end{tabular}&\begin{tabular}[c]{@{}l@{}}97.83\% \cite{el2024wireless} \end{tabular}\\ \hline

OPPT\cite{chavarriaga2013opportunity} &2013 &
\begin{tabular}[c]{@{}l@{}}Acceleration, Rate of Turn\\ Magnetic field,  Reed switches\end{tabular} & \begin{tabular}[c]{@{}l@{}}\end{tabular}
40 & 
4 & 
17 & 
\begin{tabular}[c]{@{}l@{}}Daily living activity, \\ Composite activity\end{tabular}&\begin{tabular}[c]{@{}l@{}}100\% \cite{ye2024deep}\end{tabular} \\ \hline

WISDM\cite{kwapisz2011activity} &2011 &
Accelerometer, Gyroscopes & 
1 & 
33 & 
6 & 
\begin{tabular}[c]{@{}l@{}}Daily living activity, \\ Sports fitness activity\end{tabular}&\begin{tabular}[c]{@{}l@{}} 97.8\% \cite{kaya2024human} \end{tabular} \\ \hline

UCIHAR\cite{anguita2013public} & 2013&
Accelerometer, Gyroscope & 
1 & 
30 & 
6 & 
Daily living activity &\begin{tabular}[c]{@{}l@{}} \end{tabular}\\ \hline

PAMAP2\cite{reiss2012introducing} &2012 &
\begin{tabular}[c]{@{}l@{}}Accelerometer, Gyroscope, \\ Magnetometer, \\ Temperature\end{tabular} & 
4 & 
9 & 
18 & 
\begin{tabular}[c]{@{}l@{}}Daily living activity, \\ Sports fitness activity, \\ Composite activity\end{tabular} &\begin{tabular}[c]{@{}l@{}}94.72\% \cite{zhu2023diamondnet}\\ 82.12\% \cite{ye2024deep}\\ 90.27\% \cite{kaya2024human} \end{tabular}\\ \hline

DSADS\cite{altun2010comparative} &2010 &\begin{tabular}[c]{@{}l@{}}
Accelerometer Gyroscope \\Magnetometer\end{tabular}  & 
45 & 
8 & 
19 & 
\begin{tabular}[c]{@{}l@{}}Daily living activity, \\ Sports fitness activity\end{tabular}&\begin{tabular}[c]{@{}l@{}}99.48\%\cite{zhang2024multi} \end{tabular} \\ \hline

RealWorld\cite{sztyler2016body} &2016 &
Acceleration & 
7 & 
15 & 
8 & 
\begin{tabular}[c]{@{}l@{}}Daily living activity, \\ Sports fitness activity\end{tabular}&\begin{tabular}[c]{@{}l@{}}95\% \cite{khan2024wearable} \end{tabular} \\ \hline

Exer. Activity\cite{cheng2013nuactiv} &2013 &\begin{tabular}[c]{@{}l@{}}Accelerometer, Gyroscope\end{tabular} & 
3 & 
20 & 
10 & 
\begin{tabular}[c]{@{}l@{}}Sports fitness activity\end{tabular}&\begin{tabular}[c]{@{}l@{}}- \end{tabular} \\ \hline

UTD-MHAD \cite{chen2015utd} &2015 &\begin{tabular}[c]{@{}l@{}} Accelerometer Gyroscope \\ RGB camera, depth camera \end{tabular} & 
3 & 
8 & 
27 & 
\begin{tabular}[c]{@{}l@{}}Daily living activity, \\ Sports fitness activity\\Composite activity\\ Atomic activity\end{tabular}&\begin{tabular}[c]{@{}l@{}}76.35\% \cite{zolfaghari2024sensor} \end{tabular} \\ \hline

Shoaib \cite{shoaib2014fusion} &2014 &\begin{tabular}[c]{@{}l@{}} Accelerometer Gyroscope  \end{tabular} & 
5 & 
10& 
7 & 
\begin{tabular}[c]{@{}l@{}}Daily living activity, \\ Sports fitness activity\end{tabular}&\begin{tabular}[c]{@{}l@{}}99.86\% \cite{zhang2024multii} \end{tabular} \\ \hline

TUD \cite{huynh2008discovery} &2008 &\begin{tabular}[c]{@{}l@{}} Accelerometer  \end{tabular} & 
2 & 
1& 
34 & 
\begin{tabular}[c]{@{}l@{}}Daily living activity, \\ Sports fitness\\Composite activity \end{tabular}&\begin{tabular}[c]{@{}l@{}}- \end{tabular} \\ \hline

SHAR \cite{micucci2017unimib} &2017 &\begin{tabular}[c]{@{}l@{}} Accelerometer  \end{tabular} & 
2 & 
30& 
17 & 
\begin{tabular}[c]{@{}l@{}}Daily living activity, \\ Sports fitness activity\\Atomic  activity\end{tabular}&\begin{tabular}[c]{@{}l@{}}82.79\%\cite{yao2024revisiting} \end{tabular} \\ \hline

USC-HAD \cite{zhang2012usc} &2012 &\begin{tabular}[c]{@{}l@{}} Accelerometer, Gyroscope \end{tabular} & 
1 & 
14& 
12 & 
\begin{tabular}[c]{@{}l@{}}Daily living activity, \\ Sports fitness \\activity activity\end{tabular}&\begin{tabular}[c]{@{}l@{}}97.25\%  \cite{yao2024revisiting}\end{tabular} \\ \hline

Mobi-Act \cite{vavoulas2016mobiact} &2016 &\begin{tabular}[c]{@{}l@{}} Accelerometer, Gyroscope\\orientation sensors \end{tabular} & 
1 & 
50& 
13 & 
\begin{tabular}[c]{@{}l@{}}Daily living activity,\\ Atomic activity activity\end{tabular}&\begin{tabular}[c]{@{}l@{}}75.87\% \cite{khaertdinov2023explaining} \end{tabular} \\ \hline

Motion Sense \cite{malekzadeh2018protecting} &2018 &\begin{tabular}[c]{@{}l@{}} Accelerometer, Gyroscope \end{tabular} & 
1 & 
24& 
6 & 
\begin{tabular}[c]{@{}l@{}}Daily living activity\end{tabular}&\begin{tabular}[c]{@{}l@{}}95.35\%\cite{saha2024decoding} \end{tabular} \\ \hline

van Kasteren \cite{van2011human} &2011 &\begin{tabular}[c]{@{}l@{}} switches, contacts\\ passive infrared (PIR)
 \end{tabular} & 
14 & 
1& 
10 & 
\begin{tabular}[c]{@{}l@{}}Daily living activity\\Composite activity \\activity\end{tabular}&\begin{tabular}[c]{@{}l@{}}- \end{tabular} \\ \hline

CASAS \cite{cook2012casas} &2012 &\begin{tabular}[c]{@{}l@{}} Temperature\\
Infrared motion/light sensor
 \end{tabular} & 
52 & 
1& 
7 & 
\begin{tabular}[c]{@{}l@{}}Daily living activity\\Composite activity\\ activity\end{tabular}&\begin{tabular}[c]{@{}l@{}}88.4\% \cite{kim2024clan} \end{tabular} \\ \hline

Skoda \cite{zappi2008activity} &2008 &\begin{tabular}[c]{@{}l@{}} Accelerometer
 \end{tabular} & 
19 & 
1& 
10 & 
\begin{tabular}[c]{@{}l@{}}Daily living activity\\Composite activity \\activity\end{tabular}&\begin{tabular}[c]{@{}l@{}} 97\%\cite{zhang2023human} \end{tabular} \\ \hline

Widar3.0 \cite{zheng2019zero} &2019 &\begin{tabular}[c]{@{}l@{}} Wi-Fi
 \end{tabular} & 
7 & 
1& 
6& 
\begin{tabular}[c]{@{}l@{}} Atomic activity
\end{tabular}&\begin{tabular}[c]{@{}l@{}}82.18\%\cite{chen2024aiot} \end{tabular} \\ \hline

UCI \cite{anguita2013public} &2013 &\begin{tabular}[c]{@{}l@{}} Accelerometer, Gyroscope
 \end{tabular} & 
2 & 
30& 
6& 
\begin{tabular}[c]{@{}l@{}} Human activity
\end{tabular}&\begin{tabular}[c]{@{}l@{}}95.90\% \cite{zhu2023diamondnet}\end{tabular} \\ \hline

HAPT \cite{reyes2016transition} &2016 &\begin{tabular}[c]{@{}l@{}} Accelerometer, Gyroscope
 \end{tabular} & 
1& 
30& 
12& 
\begin{tabular}[c]{@{}l@{}} Human activity
\end{tabular}&\begin{tabular}[c]{@{}l@{}}92.14\% \cite{zhu2023diamondnet}\\98.73\%\cite{zhang2024multii}\end{tabular} \\ \hline
\end{tabular} 
\end{table}

\begin{table}[!htp]
\centering
\setlength{\tabcolsep}{2pt}
\caption{Sensor data modality-based HAR models and performance. }
\label{tab:sensor_modalitiy_performance}
\begin{tabular}{|c|c|c|c|c|c|c|}
\hline
 \textbf{Author} & \textbf{Year} & \textbf{Dataset Name} & \begin{tabular}[c]{@{}c@{}}\textbf{Modality} \\ \textbf{Sensor} \\ \textbf{Name}\end{tabular} & \textbf{Methods} & \textbf{Classifier} & \begin{tabular} {@{}c@{}}\textbf{Accuracy} \\ \% \end{tabular}  \\
    \hline
Ignatov et al. \cite{ignatov2018real}&2018& \begin{tabular} {@{}c@{}}WISDM\\UCI HAR \end{tabular}&IMU Sensor&CNN&SoftMax& \begin{tabular} {@{}c@{}}  93.32\\97.63  \end{tabular}
\\ \hline

Jain et al. \cite{jain2017human}&2018& \begin{tabular} {@{}c@{}}UCI HAR \end{tabular}&IMU Sensor&Fusion based&SVM,KNN& \begin{tabular} {@{}c@{}} 97.12 \end{tabular}
\\ \hline

Chen et al. \cite{chen2019semisupervised}&2019& \begin{tabular} {@{}c@{}}MHEALTH\\PAMAP2\\UCI HAR \end{tabular}&IMU&CNN&SoftMax& \begin{tabular} {@{}c@{}}  94.05, 83.42\\81.32  \end{tabular}
\\ \hline
Alawneh et al. \cite{alawneh2020comparison}&2020& \begin{tabular} {@{}c@{}}UniMib Shar \\WISDM \end{tabular}&  \begin{tabular} {@{}c@{}}  Accelerometer\\ IMU Senso \end{tabular} r&Bi-LSTM&SoftMax& \begin{tabular} {@{}c@{}}  99.25\\98.11 \end{tabular}
\\ \hline
Lin et al. \cite{lin2020novel}&2020& \begin{tabular} {@{}c@{}}Smartwach \end{tabular}& \begin{tabular} {@{}c@{}}Accelerometer\\  gyroscope\end{tabular}&Dilated CNN&SoftMax& \begin{tabular} {@{}c@{}}  95.49 \end{tabular}
\\ \hline
Zhang et al. \cite{zhang2020data}&2020& \begin{tabular} {@{}c@{}}WaFi CSI\end{tabular}&Wafi signal&Dense-LSTM&SoftMax& \begin{tabular} {@{}c@{}}  90.0 \end{tabular}

\\ \hline

Nadeem et al. \cite{nadeem2021automatic}&2021& \begin{tabular} {@{}c@{}}WISDM\\ PAMAP2 \\ USC-HAD\end{tabular}&\begin{tabular}{@{}c@{}}IMU\end{tabular}&HMM&SoftMax& \begin{tabular} {@{}c@{}}  91.28 \\91.73\\90.19 \end{tabular}
\\ \hline

kavuncuoug et al. \cite{kavuncuouglu2022investigating}&2021& \begin{tabular} {@{}c@{}}Fall and ADLs\end{tabular}&\begin{tabular}{@{}c@{}} Accelerometer\\Gyroscope\\ Magnetometer\end{tabular}&ML& \begin{tabular} {@{}c@{}}  SVM, K-NN \end{tabular} & \begin{tabular} {@{}c@{}}  99.96 \\95.27 \end{tabular}
\\ \hline
Lu et al.  \cite{lu2022multichannel}&2022& \begin{tabular} {@{}c@{}} WISDM, PAMAP2\\UCI-HAR\\ \end{tabular}&\begin{tabular}{@{}c@{}}IMUs\\ Accelerometers\\Accelerometers\end{tabular}&CNN-GRU&SoftMax& \begin{tabular} {@{}c@{}} 96.41\\96.25\\96.67\end{tabular}
\\ \hline
Kim et al.  \cite{kim2022oversampling}&2022& \begin{tabular} {@{}c@{}}  WISDM\\ USC-HAR\end{tabular}&\begin{tabular}{@{}c@{}}IMUs\end{tabular}&CNN-BiGRU&SoftMax& \begin{tabular} {@{}c@{}}99.49\\88.31\end{tabular}
\\ \hline
Sarkar et al.  \cite{sarkar2023human}&2023& \begin{tabular} {@{}c@{}} UCI-HAR\\  WISDM, MHEALTH\\PAMAP2\\ HHAR\end{tabular}&\begin{tabular}{@{}c@{}}IMUs\\ Accelerometers\\Accelerometers\end{tabular}&CNN with GA&SVM& \begin{tabular} {@{}c@{}}  98.74\\98.34 \\99.72\\97.55\\96.87\end{tabular}
\\ \hline

Semwal et al. \cite{semwal2023gait}&2023& \begin{tabular} {@{}c@{}} WISDM\\ PAMAP2 \\ USC-HAD\end{tabular}&\begin{tabular}{@{}c@{}}IMUs\end{tabular}&CNN and LSTM&SoftMax& \begin{tabular} {@{}c@{}}  95.76\\94.64 \\89.83\end{tabular}
\\ \hline
Yao et al.  \cite{yao2024revisiting}&2024& \begin{tabular} {@{}c@{}}  PAMAP2\\ USC-HAD, UniMiB-SHAR\\ OPPORTUNITY\end{tabular}&\begin{tabular}{@{}c@{}}IMUs\\ Accelerometers\end{tabular}&ELK ResNet&SoftMax& \begin{tabular} {@{}c@{}}  95.53\\97.25\\82.79\\87.96\end{tabular}\\ \hline
Wei et al.  \cite{wei2024tcn}&2024& \begin{tabular} {@{}c@{}} WISDM\\ PAMAP2 \\ USC-HAD\end{tabular}&\begin{tabular}{@{}c@{}}IMU\end{tabular}&TCN-Attention&SoftMax& \begin{tabular} {@{}c@{}}  99.03\\98.35\\96.32\end{tabular}\\ \hline

 El‑Adawi et al. \cite{el2024wireless}&2024& \begin{tabular} {@{}c@{}}MHEALTH \end{tabular}&\begin{tabular}{@{}c@{}}IMU\end{tabular}&GAF+DenseNet169&SoftMax& \begin{tabular} {@{}c@{}} 97.83\end{tabular}\\ \hline

Ye et al. \cite{ye2024deep}&2024& \begin{tabular} {@{}c@{}}  OPPT, PAMAP2\end{tabular}&\begin{tabular}{@{}c@{}}IMU\end{tabular}& CVAE-USM&GMM& \begin{tabular} {@{}c@{}} 100\\82.12\end{tabular}\\ \hline

Kaya et al. \cite{kaya2024human}&2024& \begin{tabular} {@{}c@{}}  UCI-HAPT\\ WISDM,PAMAP2\end{tabular}&\begin{tabular}{@{}c@{}}IMU\end{tabular}& Deep CNN&SoftMax & \begin{tabular} {@{}c@{}}  98\\97.8\\90.27\end{tabular}\\ \hline

Zhang et al. \cite{zhang2024multi}&2024& \begin{tabular} {@{}c@{}} Shoaib, SisFall\\ HCIHAR, KU-HAR\end{tabular}&\begin{tabular}{@{}c@{}}IMU\end{tabular}& \begin{tabular}{@{}c@{}}1DCNN-Att\\-BiLSTM\end{tabular} &SVM & \begin{tabular} {@{}c@{}}  99.48\\ 91.85\\96.67\\  97.99\end{tabular}\\ \hline

Zhang et al. \cite{zhang2024multii}&2024& \begin{tabular} {@{}c@{}}  DSADS\\HAPT\end{tabular}&\begin{tabular}{@{}c@{}}IMU\end{tabular}& Multi-STMT&SoftMax & \begin{tabular} {@{}c@{}}  99.86\\ 98.73\end{tabular}\\ \hline

Saha et al. \cite{saha2024decoding}&2024& \begin{tabular} {@{}c@{}}   UCI HAR \\ Motion-Sense\end{tabular}&\begin{tabular}{@{}c@{}}IMU\end{tabular}& FusionActNet&SoftMax & \begin{tabular} {@{}c@{}}   97.35\\ 95.35\end{tabular}\\ \hline

\end{tabular}
\end{table}

\subsection{Preprocessing of the Sensor Dataset}
 Preprocessing sensor data is very crucial for reliable analysis and effective maintenance. Consequently, data collected from sensing devices must be preprocessed before being utilized for any analysis.  Poor data quality, including missing values, outliers, and spikes, can impact the performance results. Preprocessing steps like imputing missing data, noise reduction, and normalization are significant. A fast, scalable module is needed for real-time data preprocessing, especially in predictive maintenance systems \cite{latyshev2018sensor}. After preprocessing the sensor data, the second step is feature engineering, which involves creating new characteristics from existing data. Its main goals are to improve connections between input and output variables in forecasting models and to select the most useful features, enhancing model quality and efficiency. Finally, a proper model must be designed and implemented.   

\subsection{Sensor Data Modality Based HAR System Using Feature Extraction with Machine Learning}
Previous studies on sensor-based HAR have involved manually extracting features from raw sensor data and using conventional ML techniques like SVM, RF, KNN, DT, and NB \cite{miah2019eeg,joy2020multiclass,miah2020motor,zobaed2020real,miah2021event}. Kavuncuoglu et al. \cite{kavuncuouglu2022investigating}  combining accelerometer and magnetometer data with SVM improves fall and activity classification. Feature-level fusion has outperformed fraction-level fusion with multiclass SVM and KNN classifiers on UCI HAR and physical activity sensor datasets. Using EEG data, models like RF and GB demonstrated excellent performance \cite{jain2017human}, with LIME providing insights into significant EEG features \cite{hussain2023explainable}. Introducing new activity classifications and novel feature engineering with models like GBDT, RF, KNN, and SVM has enhanced activity recognition accuracy. However, these traditional methods depend heavily on the quality of feature engineering, requiring domain-specific expertise to extract and select relevant features, which may not generalize across all activities \cite{thakur2022convae}.

\subsection{Sensor Data Modality Based HAR System Using Deep Learning Approach}
Recently many researchers have developed DL-based methods for HAR using sensor-based datasets, such as CNNs and RNNs, which automatically learn complex features from raw sensor data without manual feature extraction. Figure \ref{fig_milestone_sensor_multimodal} demonstrates the year-wise end-to-end deep learning method developed by various researchers for sensor-based HAR systems. These models achieve state-of-the-art results HAR. However, CNNs may not capture time-domain characteristics effectively.

\subsubsection{Background of the Deep Learning Based Temporal Modeling TCN}
Recently, the study revolves around advancements in Human Activity Recognition (HAR) using ambient sensors. It highlights the integration of various types of sensors—user-driven, environment-driven, and object-driven—into HAR systems \cite{kim2024clan}. Recent progress in HAR involves leveraging DL-based techniques, including Transformer models with multi-head attention mechanisms, to effectively capture temporal dependencies in activity data \cite{wang2019deep}. Additionally, the importance of sensor frequency information and the analysis of time and frequency domains in understanding sensor-driven time series data are emphasized \cite{madsen2007time}. The previous approach performs to addresses challenges such as adapting HAR systems to new activities in dynamic environments \cite{chen2021deep}. Kim et al. \cite{kim2024clan} developed a contrastive learning-based novelty detection (CLAN) method for HAR from sensor data. They perform to address challenges like temporal and frequency features, complex activity dynamics, and sensor modality variations by leveraging diverse negative pairs through data augmentation. The two-tower model extracts invariant representations of known activities, enhancing recognition of new activities, even with shared features.  Wei et al. \cite{wei2024tcn} presented a Time Convolution Network with Attention Mechanism (TCN-Attention-HAR) model designed to enhance HAR using wearable sensor data. Addressing challenges such as effective temporal feature extraction and gradient issues in deep networks, the model optimizes feature extraction with appropriate temporal convolution sizes and prioritizes important information using attention mechanisms. 
Zhang et al. \cite{zhang2024multi} presents Multi-STMT, a multilevel model for HAR using wearable sensors that integrate spatiotemporal attention and multiscale temporal embedding; the model combines CNN and BiGRU modules with attention mechanisms to capture nuanced differences in activities.
The Conditional Variational Autoencoder with Universal Sequence Mapping (CVAE-USM) for HAR. This method addresses the challenge of non-i.i.d. data distributions in cross-user scenarios by leveraging temporal relationships in time-series data. They combining VAE and USM techniques, CVAE-USM effectively aligns user data distributions, capturing common temporal patterns to enhance activity recognition accuracy.
\subsubsection{CNN based Various Stream for HAR}

Ignatov et al. \cite{ignatov2018real} utilized a DL-based approach for real-time HAR with mobile sensor data. They employ CNN for local feature extraction and integrate simple statistical features to capture global time series patterns. The experimental evaluations of the WISDM and UCI datasets demonstrate high accuracy across various users and datasets, highlighting their effectiveness in the DL-based method without needing complex computational resources or manual feature engineering. Chen et al. \cite{chen2019semisupervised} developed a semi-supervised DL-based model for imbalanced HAR utilized multimodal wearable sensory data. Addressing challenges such as limited labelled data and class imbalance, the model employs a pattern-balanced framework to extract diverse activity patterns. They used recurrent convolutional attention networks to identify salient features across modalities. Kaya et al. \cite{kaya2024human} presented a 1D-CNN-based approach to accurately HAR from sensor data. They evaluated their model using raw accelerometer and gyroscope sensor data from three public datasets: UCI-HAPT, WISDM, and PAMAP2. Zhang et al. \cite{zhang2023human} presented a method HAR using sensor data modality called ConvTransformer. They combine CNN, Transformer, and attention mechanisms to handle the challenge of extracting both detailed and overall features from sensor data.
\subsubsection{RNN, LSTM, Bi-LSTM for HAR}
In most of the recent work, including RNNs \cite{ordonez2016deep} play a crucial role in handling temporal dependencies in sensor data for HAR. To address challenges like gradient issues, LSTM networks were developed \cite{murad2017deep}. Researchers \cite{alawneh2020comparison,gupta2021deep,zhang2024multi,zhang2020data} have also explored attention-based BiLSTM models, achieving the best performance compared to other DL-based methods. The experimental evaluations on various datasets shown in Table \ref{tab:sensor_modalitiy_performance} demonstrate high accuracy across various users and datasets, highlighting their effectiveness in the DL-based method without needing complex computational resources or manual feature engineering. Saha et al. \cite{saha2024decoding} presented  Fusion ActNet, an advanced method for HAR using sensor data. It features dedicated residual networks to capture static and dynamic actions separately, alongside a guidance module for decision-making, through a two-stage training process and evaluations on benchmark datasets.  Murad et al. \cite{murad2017deep} used deep recurrent neural networks (DRNNs) in HAR, highlighting their ability to capture long-range dependencies in variable-length input sequences from body-worn sensors. Unlike traditional approaches that overlook temporal correlations, DRNNs, including unidirectional, bidirectional, and cascaded LSTM frameworks, perform well on diverse benchmark datasets. They perform the comparison of conventional machine learning approaches like SVM and KNN, as well as other deep learning techniques such as DBNs and CNNs, demonstrating their effectiveness in activity recognition tasks. 

\subsubsection{Integratation CNN and LSTM Based Technique}
Several studies have developed that utilize hybrid models, combining different DL architectures can report high-performance accuracy in HAR. For instance, a hybrid CNN-LSTM model \cite{chen2020attention,semwal2023gait} improved sleep-wake detection using heterogeneous sensors. Additionally, designs like TCCSNet \cite{essa2023temporal} and CSNet leverage temporal and channel dependencies to enhance human behaviour detection.  Ordonez et al. \cite{ordonez2016deep} developed a model for  HAR using CNN and LSTM recurrent units. They extract features from raw sensor data, support multimodal sensor fusion, and model complex temporal dynamics without manual feature design. Evaluation of benchmark datasets, such as Opportunity and Skoda, shows significant performance improvements over traditional methods, highlighting their effectiveness in HAR applications. Zhang et al \cite{zhang2020multi} developed a multi-channel DL-based network called a hybrid model (1DCNN-Att-BiLSTM) for improved recognition performance, evaluation using publicly accessible datasets, and comparison with ML and DL models. El-adawi et al. \cite{el2024wireless} developed a HAR model within a Wireless Body Area Network (WBAN). The model leverages the Gramian Angular Field (GAF) and DenseNet. By converting time series data into 2D images using GAF and integrating them with DenseNet they achieve good performance accuracy.

\section{Multimodal Fusion Modality Based Action Recognition}
\label{section_5}
Actions recognition through the utilization of a dataset that consists of multiple modalities necessitates the act of discerning and categorizing human actions or activities. This dataset encompasses various forms of information, including visual, audio, and sensor data. Integrating diverse sources of information within multi-modal datasets affords a better comprehension of actions. From the perspective of the input data's modality, DL techniques can acquire human action characteristics through a diverse range of modal data. Similarly, the ML-based algorithm aims to process the information from multiple modalities. By using the strengths of various data types,  multi-modal ML can often perform more accurate HAR tasks. There are several types of multimodality learning methods, including fusion-based methods such as RGB with skeleton and dept-based modalities. Generally, fusion refers to combining the information of two or more modalities to train the model and provide accurate results of HAR. There are two main approaches widely utilized in multi-modality fusion schemes, namely score fusion and feature fusion. The fusion-based approach combines scores obtained from various sources, including weight averaging \cite{rani2021kinematic} or learning a score fusion \cite{dhiman2020view} model, while the feature fusion \cite{wang2019generative} focuses on integrating features extracted from different modalities. Ramani et al. \cite{rahmani2014real}  developed an algorithm that combines depth image and 3D joint position data using local spatiotemporal features and dominant skeleton movements. 
Researchers have increasingly explored DL techniques to extract action-effective features utilizing the RGB, depth, and skeleton data. These methods facilitate multimodal feature learning from deep networks \cite{simonyan2014two,tran2015learning,liu2016spatio,wang2015action}, encompassing appearance image information such as optical flow sequences, depth sequences, and skeleton sequences. DL networks are proficient at learning human action effective features by performing single-modal data or multimodal fusion data \cite{guler2018densepose,fang2017rmpe,cao2017realtime}. Note that score fusion and feature fusion are important in advancing HAR technology to provide accurate results.
Table \ref{Table_multi_modality_Analysis} lists the basic information of the existing model, including datasets, multi-modality, features extraction methods, classifier, years, and performance accuracy.\\

\subsection{Multimodal Fusion Based HAR Dataset}
We provided the most popular benchmark HAR datasets, which come from the multi-modal fusion dataset, which is demonstrated in Table \ref{Table_multi_modality_Analysis}. The dataset table demonstrated the details of the datasets, including modalities, creation year, number of classes, number of subjects who participated in recording the dataset, number of samples, and latest performance accuracy of the dataset with citation. Figure \ref{fig_milestone_sensor_multimodal} also demonstrates the year-wise end-to-end deep learning method developed by various researchers for multimodal fusion-based HAR systems. 

Table \ref{tab:multi_modalities} presents a comprehensive overview of benchmark datasets for Human Activity Recognition (HAR) using various modalities. The datasets include combinations of RGB, Skeleton, Depth, Infrared, Acceleration, and Gyroscope data, providing rich and diverse sources for model training and evaluation. For instance, the MSRDailyActivity3D dataset, introduced in 2012, includes RGB, Skeleton, and Depth data, featuring 16 classes, 10 subjects, and 320 samples with a notable accuracy of 97.50\% \cite{wang2012mining} \cite{shahroudy2017deep}. The N-UCLA dataset from 2014 also incorporates RGB, Skeleton, and Depth data, spanning 10 classes, 10 subjects, and 1475 samples, achieving an impressive 99.10\% accuracy \cite{wang2014cross} \cite{cheng2024dense}. Another significant dataset, NTU RGB+D, initially released in 2016 and updated in 2019, includes RGB, Skeleton, Depth, and Infrared modalities, with 60 and 120 classes, 40 and 106 subjects, and 56880 and 114480 samples respectively, both recording a high accuracy of 97.40\% \cite{shahroudy2016ntu} \cite{liu2019ntu} \cite{cheng2024dense}. The Kinetics-600 dataset, published in 2018, is one of the largest, containing RGB, Skeleton, Depth, and Infrared data across 600 classes and 595445 samples, with an accuracy of 91.90\% \cite{carreira2018short} \cite{wang2024internvideo2}. These datasets are crucial for advancing HAR research, offering extensive and varied data for developing robust and accurate models.

\begin{table}[]
\centering
\setlength{\tabcolsep}{2pt}
\caption{Multimodality fusion based HAR benchmark datasets.}
\label{tab:multi_modalities}
\begin{tabular}{|c|c|c|c|c|c|c|c|}
\hline
\textbf{Dataset} & \textbf{Data set modalities} & \textbf{Year} & \textbf{Class} & \textbf{Subject} & \textbf{Sample} & \textbf{Latest Accuracy} \\ \hline
\begin{tabular}[c]{@{}c@{}}MSRDaily\\Activity3D \cite{wang2012mining}\end{tabular} & RGB, Skeleton, Depth & 2012 & 16 & 10 & 320 & 97.50\% \cite{shahroudy2017deep} \\ \hline
\begin{tabular}[c]{@{}c@{}}N-UCLA \cite{wang2014cross}\end{tabular} & RGB, Skeleton, Depth & 2014 & 10 & 10 & 1475 & 99.10\% \cite{cheng2024dense} \\ \hline
\begin{tabular}[c]{@{}c@{}}Multi-View TJU \cite{liu2014multiple}\end{tabular} & RGB, Skeleton, Depth & 2014 & 20 & 22 & 7040 & - \\ \hline
\begin{tabular}[c]{@{}c@{}}UTD-MHAD \cite{chen2015utd}\end{tabular} & RGB, Skeleton, Depth, Acceleration, Gyroscope & 2015 & 27 & 8 & 861 & 95.0\% \cite{liu2018recognizing} \\ \hline
\begin{tabular}[c]{@{}c@{}}UWA3D\\Multiview II  \cite{rahmani2016histogram}\end{tabular} & RGB, Skeleton, Depth & 2015 & 30 & 10 & 1075 & - \\ \hline
\begin{tabular}[c]{@{}c@{}}NTU RGB+D \cite{shahroudy2016ntu}\end{tabular} & RGB, Skeleton, Depth, Infrared & 2016 & 60 & 40 & 56880 & 97.40\% \cite{cheng2024dense} \\ \hline
\begin{tabular}[c]{@{}c@{}}PKU-MMD \cite{liu2017pku}\end{tabular} & RGB, Skeleton, Depth, Infrared & 2017 & 51 & 66 & 10076 & 94.40\% \cite{li2019making} \\ \hline
\begin{tabular}[c]{@{}c@{}}NEU-UB \cite{kong2017max}\end{tabular} & RGB, Depth & 2017 & 6 & 20 & 600 & - \\ \hline
\begin{tabular}[c]{@{}c@{}}Kinetics-600 \cite{carreira2018short}\end{tabular} & RGB, Skeleton, Depth, Infrared & 2018 & 600 & - & 595445 & 91.90\% \cite{wang2024internvideo2} \\ \hline
\begin{tabular}[c]{@{}c@{}}RGB-D \\Varing-View \cite{ji2018large}\end{tabular} & RGB, Skeleton, Depth & 2018 & 40 & 118 & 25600 & - \\ \hline
\begin{tabular}[c]{@{}c@{}}Drive\&Act \cite{martin2019drive}\end{tabular} & RGB, Skeleton, Depth & 2019 & 83 & 15 & - & 77.61\% \cite{lin2024multi} \\ \hline
\begin{tabular}[c]{@{}c@{}}MMAct \cite{kong2019mmact}\end{tabular} & RGB, Skeleton, Acceleration, Gyroscope & 2019 & 37 & 20 & 36764 & 98.60\% \cite{liu2021semantics} \\ \hline
\begin{tabular}[c]{@{}c@{}}Toyota-SH \cite{das2019toyota}\end{tabular} & RGB, Skeleton, Depth & 2019 & 31 & 18 & 16115 & - \\ \hline
\begin{tabular}[c]{@{}c@{}}IKEA ASM \cite{ben2021ikea}\end{tabular} & RGB, Skeleton, Depth & 2020 & 33 & 48 & 16764 & - \\ \hline
\begin{tabular}[c]{@{}c@{}}ETRI-Activity3D \cite{jang2020etri}\end{tabular} & RGB, Skeleton, Depth & 2020 & 55 & 100 & 112620 & 95.09\% \cite{dokkar2023convivit} \\ \hline
\begin{tabular}[c]{@{}c@{}}UAV-Human \cite{li2021uav}\end{tabular} & RGB, Skeleton, Depth & 2021 & 155 & 119 & 27428 & 55.00\% \cite{xian2024pmi} \\ \hline
\end{tabular}
\end{table}

\subsection{Fusion of RGB, Skeleton, and Depth Modalities} 
Recently, several hand-crafted feature-based approaches \cite{kong2017max} \cite{hu2015jointly} have been developed to explore multi-modalities such as RGB, skeleton, and depth to improve the performance of the action recognition tasks. While the DL-based approaches \cite{shahroudy2017deep,hu2018deep,khaire2018combining,cardenas2018multimodal} have been proposed due to providing good performance. Shahoudy et al. \cite{shahroudy2017deep} study and explore the concept of correlation analysis between the different modalities and factorize them into desired independent components. They used a structured spared classifier for the HAR task.  Hu et al. \cite{hu2018deep} analysis between the time-varying information across the fusion of multimodality such as RGB, Skelton, and depth-based. They extracted temporal features from each modality and then concatenated them along the desired modality dimension. These multi-modal temporal features were then input into the model. Khaire et al. \cite{khaire2018combining} developed a CNN network with five streams. These streams take inputs from MHI \cite{bobick2001recognition}, DMM \cite{yang2012recognizing}, and skeleton images generated from RGB, depth, and skeleton sequences. Each CNN stream was trained separately, and the final classification scores were obtained by combining the output scores of all five CNN streams utilizing a weighted product model. Similarly, Khair et al. \cite{khaire2018human}, a fusion of three methods to merge skeletal, RGB, and depth modalities. Cardens et al. \cite{cardenas2018multimodal} utilized three distinct optical spectra channels from skeleton data \cite{hou2016skeleton} and dynamic images from RGB and depth videos. These features were fed into a pre-trained CNN to extract multi-modal features. Finally, they used a feature aggregation module for classification tasks.
\subsection{Fusion of Signal and Visual Modalities}
Signal data complements visual data by providing additional information. Various DL-based approaches have been proposed to merge these modalities for HAR. Wang et al. \cite{wang2016exploring} proposed three-stream CNN models to extract features from multimodalities. They evaluated the performance of both feature fusion and score fusion, with feature fusion showing superior performance. Owens et al. proposed a model of a two-stream CNN in a self-supervised manner to detect misalignments between audio and visual sequences.
Subsequently, they refined the model using HAR datasets for audio-visual recognition. TSN \cite{wang2016temporal} showed improved performance by Kazakos et al. \cite{kazakos2019epic} introduced the Temporal Binding Network (TBN) for egocentric HAR, integrating audio, RGB, and optical flow inputs. TBN utilized a three-stream CNN to merge these inputs within each Temporal Binding Window, enhancing classification through temporal aggregation. Their findings demonstrated TBN's superiority over TSN \cite{wang2016temporal} in audio-visual HAR tasks. Additionally, Gao et al. \cite{gao2020listen} utilized audio data to minimize temporal redundancies in videos, employing knowledge distillation from a teacher network trained on video clips to a student network trained on image-audio pairs for efficient HAR. Xiao et al. \cite{xiao2020audiovisual} developed a novel framework combining audio and visual information, incorporating slow and fast visual pathways alongside a faster audio pathway across multiple layers. They employed two training strategies: randomly dropping the audio pathway and hierarchical audio-visual synchronization, facilitating the training of audio-video integration.
In addition, the multimodal HAR-based approaches such as
Bruce et al. \cite{bruce2022mmnet} multimodal network (MMNet) fuses skeleton and RGB data using a spatiotemporal GNN to transfer attention weights, significantly improving HAR accuracy while Venkatachalam et al. \cite{venkatachalam2023bimodal} proposes a hybrid 1D CNN with LSTM classifier for HAR.
Overall, the objective of data fusion methods is to capitalize on the benefits of integrating various datasets to achieve a more robust and comprehensive feature representation. Consequently, the central issue that arises in the development of most data-fusion-based techniques revolves around determining the most efficient manner in which to combine disparate data types. This is typically addressed by employing the conventional early and late fusion strategies. The initial fusion occurs at the feature level, involving feature concatenation as the input to the recognition model. In contrast, the latter scenario performs fusion at the score level, integrating the output scores of the recognition model with diverse data types. The multimodal data fusion methods generally yield better recognition results than single-data approaches. However, the multimodal data fusion methods approach requires processing larger datasets and dealing with higher feature dimensions, thereby increasing the computational complexity of action recognition algorithms.

\begin{table}[!htp] 
\centering
\setlength{\tabcolsep}{2pt}
\caption{Multi modality data fusion based HAR system models and performance.}
\label{Table_multi_modality_Analysis}
\begin{tabular}{|c|c|c|c|c|c|c|}
\hline
\textbf{Dataset}& \textbf{Classifier}&\textbf{Methods} & \textbf{Data set type}  & \textbf{Year} & \textbf{Reference} &\textbf{Accuracy[\%]}\\ \hline
\begin{tabular} {@{}c@{}}
 NTU RGB+D (CS)\\NTU RGB+D (CV)\end{tabular}&SVM&P-LSTM& RGB, Depth& 2016  &\cite{shahroudy2016ntu} & \begin{tabular} {@{}c@{}} 62.93\\70.27 \end{tabular}
\\ \hline 
\begin{tabular} {@{}c@{}}UCI-HAD\\USC-HAD \\Opportunity\\Daphnet FOG\\Skoda\end{tabular}&SVM  KNN&DRNN & Sensors & 2017  &\cite{murad2017deep} & \begin{tabular} {@{}c@{}} 96.7\\97.8\\92.5\\94.1\\92.6 \end{tabular}
\\ \hline 
 \begin{tabular} {@{}c@{}}Smartwach \end{tabular}&SoftMax&Dilated CNN& sensor&2020 & \cite{lin2020novel} & \begin{tabular} {@{}c@{}}95.49\end{tabular}
\\ \hline
\begin{tabular} {@{}c@{}} UTD-MHAD \\ NTU RGB+D\end{tabular}&SoftMax&Vission based&RGB,Depth,Skeleton&2021 & \cite{romaissa2021vision} & \begin{tabular} {@{}c@{}}98.88\\75.50\end{tabular}
\\ \hline
 \begin{tabular} {@{}c@{}}NTU RGB+D (CS)\\NTU RGB+D (CV) \\ SYSU 3D HOI \\ UWA3D II\end{tabular}& \begin{tabular} {@{}c@{}} hierarchical-\\score fusion\end{tabular} &Multi Model & RGB Depth&2021 & \cite{ren2021multi} & \begin{tabular} {@{}c@{}}89.70\\ 92.97 \\87.08\end{tabular}
\\ \hline
  \begin{tabular} {@{}c@{}}UCF-101 \\Something-Something-v2\\  Kinetics-600\end{tabular}&SoftMax&MM-ViT & RGB &2022  & \cite{chen2022mm} & \begin{tabular} {@{}c@{}}98.9 \\90.8\\96.8\end{tabular}
\\ \hline
\begin{tabular} {@{}c@{}}MHEALTH \\UCI-HAR\end{tabular}&SoftMax&CNN-LSTM & Sensor &2022  & \cite{khatun2022deep} & \begin{tabular} {@{}c@{}}98.76 \\93.11 \end{tabular}
\\ \hline
\begin{tabular} {@{}c@{}} UCI-HAR\\  WISDM \\MHEALTH\\PAMAP2\\ HHAR\end{tabular}&SVM& CNN with GA &Sensors&2023 &\cite{sarkar2023human}& \begin{tabular} {@{}c@{}} 98.74\\98.34 \\99.72\\97.55\\96.87\end{tabular}
\\ \hline
\begin{tabular} {@{}c@{}} NTU RGB+D 60\\ NTU RGB+D120\\PKU-MMD\\Northwestern\\UCLAMultiview\\Toyota Smarthome\end{tabular}&-&MMNet &RGB, Depth&2023 &\cite{bruce2022mmnet}& \begin{tabular} {@{}c@{}} 98.0\\90.5\\98.0\\93.3\end{tabular}
\\ \hline
\begin{tabular} {@{}c@{}} NTU RGB+D 60\\ NTU RGB+D120\\ NW-UCLA\end{tabular}&SoftMax& InfoGCN &RGB, Depth&2023 &\cite{chi2022infogcn}& \begin{tabular} {@{}c@{}}93.0\\ 89.8\\97.0\end{tabular}
\\ \hline
\begin{tabular} {@{}c@{}} NTU RGB+D  \\ NTU RGB+D120\end{tabular}&Softmax&Two-stream Transformer &RGB, Depth&2023 &\cite{wang20233mformer}& \begin{tabular} {@{}c@{}} 94.8\\93.8\end{tabular}
\\ \hline
\begin{tabular} {@{}c@{}} NTU RGB+D \\ NTU RGB+D120\\NW-UCLA \end{tabular}&Softmax&  \begin{tabular} {@{}c@{}} Language\\ knowledge-assisted\end{tabular}  &RGB, Depth&2023 &\cite{xu2023language}& \begin{tabular} {@{}c@{}} 97.2\\91.8\\97.6\end{tabular}
\\ \hline
\begin{tabular} {@{}c@{}} UCF51\\Kinetics Sound\end{tabular}&SoftMax&MAIVAR-CH &RGB, audio&2024 &\cite{shaikh2024multimodal}& \begin{tabular} {@{}c@{}} 87.9\\79.0\end{tabular}
\\ \hline
\begin{tabular} {@{}c@{}}Drive Act\end{tabular}&-&Dual Feature Shift &RGB, Depth, Infrared&2024 &\cite{lin2024multi}& \begin{tabular} {@{}c@{}} 77.61\end{tabular}
\\ \hline
\begin{tabular}{@{}c@{}}Florence3DAction \\ UTKinect-Action3D \\ 3DActionPairs \\ NTURGB+D\end{tabular} &
    SoftMax &
    \begin{tabular}{@{}c@{}}two-stream \\ spatial-temporal \\ architecture\end{tabular} &
    RGB, Depth, Infrared &
    2024 &
    \cite{dai2024light} &
    \begin{tabular}{@{}c@{}}93.8 \\ 98.7 \\ 97.3 \\ 90.2\end{tabular}    
\\ \hline
\end{tabular}

\end{table}

\section{Current Challenges} \label{section_6}
Although notable progress in HAR utilizing four data modalities, several challenges persist due to the intricate nature of the various aspects of this task.
\subsection{RGB Data Modality Based Current Challenges}
The researcher explores the challenges specific to RGB-based methods in HAR. RGB data, which represents color information from regular images or videos, is widely used for determining human actions. In the following section, we described key challenges associated with RGB-based HAR:

\subsubsection{Efficient Action Recognition Analysis}
The good performance of numerous HAR approaches often comes with the cost of high computational complexity. However, an efficient HAR system is vital for many real-world applications. Therefore, it is essential to explore ways to minimize computational costs (such as CPU, GPU, and energy usage) to perform efficient and fast HAR. These limitations led to a notable impact on the computation efficiency of the network. Additionally, the process of accurately and efficiently labeling video data incurs substantial labor and time expenses due to the diversity and scale of the data.

\subsubsection{Complexity within the Environment}
Certain HAR techniques perform strongly in controlled environments but tend to underperform in uncontrolled outdoor settings. This is mostly caused by motion vector noise, which can drastically degrade resolution. Extracting effective features from complex images is an extremely tough task. For example, the rapid movement of the camera complicates the extraction of effective action features. Accurate feature extraction will also affect environmental issues such as (poor lighting, dynamic background, etc.)
\subsubsection{Large Memory of the Dataset and Limitations }
The dataset exhibits both intra-class variation and inter-class similarity. Many people perform the same action in diverse manners, and even a single person may execute it in multiple ways. Additionally,  different actions might have similar presentations. Furthermore, many existing datasets include unfiltered sequences, potentially compromising the timeliness and reducing the HAR accuracy of the model. \\
The dataset's large memory requirements pose significant limitations, particularly in terms of storage and processing capabilities. Handling massive amounts of data necessitates robust computational resources, including high-capacity storage solutions and powerful processing units. Additionally, working with large datasets may lead to challenges related to data transfer speeds, memory management, and computational efficiency. These limitations can impact the scalability, accessibility, and usability of the dataset, potentially hindering its widespread adoption and utilization in research and applications. Therefore, addressing the constraints posed by the dataset's large memory footprint is crucial for maximizing its utility and effectiveness in various domains.
\subsection{Skeleton Data Modality Based Challenges}
The challenges are specific to skeleton-based approaches in HAR. Skeleton data, which obtained joint positions and movements, is a valuable modality for understanding human actions. In the following section, some key challenges are described.
\subsubsection{Pose Preparation and Analysis}
Depending on depth cameras and sensors, Skeleton data acquisition is affected by environmental complexity, capture duration, and equipment exposure conditions. Another common challenge in daily life scenarios is Occlusion, caused by surrounding objects or human interaction, which further contributes to detection errors in skeletons.
\subsubsection{Viewpoint Variation}
Accurately distinguishing skeleton features from different perspectives poses a significant challenge, as certain features may be lost during changes in viewpoint. While modern RGBD cameras \cite{keselman2017intel,drouin2020consumer,grunnet2018best,zabatani2019intel} can normalize 3D human skeletons \cite{yang2014effective,pazhoumand2015joint} from various angles to a single pose with viewpoint invariance utilizing pose estimation transformation matrices. However, in this process, there is a risk of losing some of the relative motion between the original skeletons. This loss of relative motion can impact the accuracy and completeness of the skeleton data, highlighting the need for careful consideration and validation of viewpoint normalization techniques in skeleton feature extraction.

\subsubsection{Single Scale Data Analysis}  
As several skeleton-based datasets mostly provide information based on the scale of body joints, numerous techniques focus solely on extracting features related to the human joint scale. However, this technique often leads to the loss of fine joint features. Moreover, certain actions, such as shaving, tooth brushing, and applying lipstick, exhibit similar joint interactions. Therefore, there is a critical need to enhance local feature extraction while maintaining the effectiveness of holistic feature extraction techniques \cite{li2020multi,parsa2020spatio,zhu2020topology,li2021symbiotic}. This improvement is crucial for achieving more accurate action recognition and understanding subtle variations in human movements.
Even though DL methods yield superior recognition performance compared to handcrafted action features, certain challenges persist in recognizing human actions based on DL, particularly in the fusion of multimodal data in DL methods. Most of the aforementioned DL-based approaches concentrate on learning action features from diverse modality data; however, only a few studies address the fusion of multimodal data. The effective fusion based on multimodal data: (RGB, optical flow, depth, and skeleton data) remains a significant unresolved challenge in HAR and DL. This area also represents a prominent research focus within HAR.

\subsection{Sensor Based HAR Current Challenges and Possible Solution}
In sensor-based HAR, different activities with similar characteristics (like walking and running) pose a challenge for feature extraction. Creating unique features to represent each activity becomes difficult due to this inter-activity similarity.\\
Another challenge is annotation scarcity due to expensive data collection and class imbalance, particularly for rare or unexpected activities.
In sensor-based HAR, three critical factors—users, time, and sensors—contribute to distribution discrepancies between training and test data. These factors include person-dependent activity patterns, evolving activity concepts over time, and diverse sensor configurations.
 When designing a HAR system, two key considerations are resource efficiency for portable devices and addressing privacy risks associated with continuous life recording. \\
  When dealing with sensory data, accurate recognition solutions must address interpretability and understand which parts of the data contribute to recognition and which parts introduce noise.
\subsection{Multimodal-Based Challenges}
In the field of HAR, researchers have explored many multi-modality approaches. These approaches include multi-modality fusion-based and cross-modality transfer learning. The fusion of data from different modalities, which can often complement each other, leads to enhancing HAR accuracy. However, it’s important to note that several existing multi-modality approaches are not as effective due to some challenges, such as overfitting, missing modalities, heterogeneous data modalities, and temporal synchronization. These suggestions that there are still possibilities to develop more effective fusion systems for multi-modality HAR.
\section{Discussion}\label{section_7}
 We describe several potential directions for future research by amalgamating the current state of affairs and addressing the methodological and application-related challenges in RGB-based,skeleton-based, sensor modality-based, and multimodal-based HAR.

\subsection{Development of the New Large Scale Datasets}
Data is as very essential to DL as model construction. However, existing datasets pose challenges when it comes to generalizing to realistic scenes. Factors like realistic surroundings and dataset size play an important role in this complexity. Additionally, most of the datasets are mainly focused on spatial representation \cite{li2018resound}. Unfortunately, there’s a scarcity of long-term modeling datasets. A notable issue arises due to regional constraints and privacy concerns. YouTube dataset managers commonly provide only video IDs or links for download rather than the actual video content. Consequently, some videos become inaccessible over time, resulting in an annual loss of approximately 5\% of videos \cite{zhu2020comprehensive}. To address these difficulties, researchers are actively working on gathering fresh datasets. These new datasets will contribute to advancing DL research and improving model performance.
\subsection{Data Augmentation Techniques}
Deep neural networks exhibit exceptional performance when trained on diverse datasets. However, limited data availability remains still a challenge. To overcome this issue, data augmentation plays an important role. In the domain of image recognition, various augmentation techniques have been proposed, spanning both DL-based techniques and simple image-processing approaches.

These approaches include random erasing \cite{zhong2020random} , generative adversarial networks (GANs) \cite{bowles2018gan}, kernel filters \cite{kang2017patchshuffle}, feature space augmentation \cite{devries2017dataset}, adversarial training \cite{li2018learning}, generative adversarial networks (GANs) \cite{bowles2018gan}, and meta-learning \cite{real2017large}.
For HAR, typical data augmentation techniques involve horizontal flipping, subclip extraction, and video merging \cite{zou2023learning}. However, these generated videos often lack realism. To overcome this limitation, Zhang et al. \cite{zhang2020self} used GANs to generate new data samples and implemented a 'self-paced selection' strategy during training. Meanwhile, Gowda et al. \cite{gowda2022learn2augment} introduced Learn2Augment, which synthesizes videos from foreground and background sources as a method for data augmentation, resulting in diverse and realistic samples.

\subsection{Advancements in Models Performances}
HAR research predominantly revolves around DL-based models, much like other advancements in computer vision. Presently, ongoing progress in deep architectures is important for HAR including the RGB-based, skeleton-based, and multimodal-based approaches to perform the action recognition task. These advancements typically focus on the following key areas of model improvement.

\begin{itemize}
    \item Long-term Dependency Analysis: 
    Long-term correlations refer to the unfolding sequence of actions that occur over extended periods, akin to how memories are stored in our brains. When we reminisce about an event, one pattern naturally triggers the next. In the context of action recognition, it is important to focus not only on spatial modeling but also on the temporal component. This emphasis arises from the remarkably strong correlations observed between adjacent temporal features.

    \item Multimodal Modeling: 
    This involves integrating data from various devices, such as audiovisual sensors. There are two primary approaches to multi-modality video understanding.

    \item Enhancing Video Representations: 
    The multi-modality data (such as depth, skeleton, and RGB information) is used to improve video representations \cite{gabeur2020multi,piergiovanni2020learning}. These representations can include scene understanding, object recognition, action detection, and audio analysis, using multimodality data like RGB, skeleton, and depth.
\end{itemize}

\begin{itemize}
    \item Efficient Modeling Analysis: 
    Creating an efficient network architecture is crucial due to the challenges posed by existing models, including model complexity, excessive parameters, and real-time performance limitations. To address these issues, techniques like distributed training \cite{lin2019training}, mobile networks \cite{howard2019searching}, hybrid precision training, model compression, quantization, and pruning can be explored. These approaches can enhance both efficiency and effectiveness in image classification tasks.

    \item Semi-supervised and Unsupervised Learning Approaches: 
    Supervised learning approaches, especially those based on deep learning, typically require large, expensive labeled datasets for model training. In contrast, unsupervised and semi-supervised learning techniques \cite{singh2021semi} can utilize unlabeled data to train models, thereby reducing the need for extensive labeled datasets. Given that unlabeled action samples are often easier to collect, unsupervised and semi-supervised approaches to Human Activity Recognition (HAR) represent a crucial research direction deserving further exploration.
\end{itemize}

\subsection{Video Lengths in Human Action Recognition}    
The action prediction tasks can be broadly categorized into short-term and long-term predictions. Short-term prediction involves predicting action labels from partially observed actions, typically seen in short videos lasting a few seconds. In contrast, long-term prediction assumes that current actions influence future actions and focuses on longer videos spanning several minutes, simulating changes in actions over time. Formally, given an action video \( x_a \), which may depict either a complete or incomplete action sequence, the objective is to predict the subsequent action \( x_b \). These actions, \( x_a \) and \( x_b \), are independent yet semantically significant, with a temporal relationship \cite{yu1806human}.
To advance action prediction research, it is essential to discover and model temporal correlations within vast datasets. Unexplored directions include understanding interpretability across different time scales, devising effective methods for modeling long-term dependencies, and leveraging multimodal data to enhance predictive models.

\subsection{Limitations}
This study was focused on research papers published between 2014 and 2024, exclusively in English, excluding relevant studies in other languages. We exclusively considered studies that utilized visual data, including  HAR feature ML-based and DL-based methods involving different data types, including RGB handcrafted features and DL-based action recognition RGB and skeleton-based methods for multimodal datasets such as RGB, depth, and skeleton, excluding on EMG based data. Furthermore, the diverse input methods and dataset variations across reviewed studies hindered direct result comparisons. Notably, some articles lacked statistical confidence intervals, making it challenging to compare their findings.
\section{Conclusion}
\label{section_8}
HAR is an important task among multiple domains within the field of computer vision, including human-computer interaction, robotics, surveillance, and security. In the past decades, it has necessitated the proficient comprehension and interpretation of human actions with various data modalities. Researchers still find the HAR task challenging in real scenes due to various complicating factors in different data modalities, including various body positions, motions, and complex background occlusion. In the study, we presented a comprehensive survey of HAR methods, including advancements across various data modalities. We briefly reviewed human action recognition techniques, including hand-crafted features in RGB, skeleton, sensor, and multi-modality fusion with conventional and end-to-end DL-based action representation techniques. Moreover, we have also reviewed the most popular benchmark datasets of the RGB, skeleton, sensor, and fusion-based modalities with the latest performance accuracy. After providing an overview of the literature about each research direction in human activity recognition, the primary effective techniques were presented to familiarize researchers with the relevant research domains. The fundamental findings of this investigation on the study of human action recognition are summarized to help researchers, especially in the field of HAR.

\section*{Author contributions}
All authors contributed equally to this work.

\section*{Funding} This work was supported by the Competitive Research Fund of The University of Aizu, Japan.
\section*{Data Availability} The data used to support the findings of this study are included in the article
\section*{Conflict of interest}
The authors declare no conflict of interest.
\bibliography{sn-bibliography}
   
\end{document}